\documentclass[11pt]{article}
\usepackage[margin=1.05in]{geometry}
\usepackage[T1]{fontenc}
\usepackage{amsmath}
\usepackage{amssymb}
\usepackage{amsthm}
\usepackage{graphicx}
\usepackage{booktabs}
\usepackage[table]{xcolor}
\usepackage{colortbl}
\usepackage{appendix}
\usepackage{algorithm}
\usepackage[noend]{algpseudocode}
\usepackage[colorlinks=true, linkcolor=blue!60!black, citecolor=blue!60!black, urlcolor=blue!60!black]{hyperref}

\definecolor{rstaclr}{RGB}{214,232,245}
\newcommand{\tabhead}[1]{\textbf{#1}}
\newcommand{\Tabind}{}
\newcommand{\colrule}{\midrule}
\newcommand{\botrule}{\bottomrule}
\newcommand{\tbl}[2]{\caption{#1}\smallskip\centering{\footnotesize\setlength{\tabcolsep}{5pt}#2}}
\newcommand{\linkref}[2][]{eqn~(\ref{#2})}

\newcommand{\digitbox}[1]{\mbox{\fboxsep=1pt  \fbox{\parbox[c][0.75em][c]{0.75em}{\centering\ensuremath{#1}}}}}
\newcommand{\mnistimg}[1]{%
  \raisebox{-0.15em}{\includegraphics[height=1.05em,trim=1 1 1 1,clip]{figures/MNIST_Digits/#1.png}}%
}

\newtheorem{theorem}{Theorem}[section]

\newtheorem{corollary}{Corollary}[section]
\newtheorem{lemma}[theorem]{Lemma}
\newtheorem{proposition}[theorem]{Proposition}
\newtheorem{definition}[theorem]{Definition}
\numberwithin{equation}{section}
\renewcommand{\thesubsection}{\thesection\alph{subsection}}

\title{Neuro-Symbolic Weak Supervision: Theory and Semantics}
\author{Nijesh Upreti\qquad Vaishak Belle\\[4pt]
\normalsize School of Informatics, University of Edinburgh, Edinburgh, UK}
\date{}

\begin{document}
\maketitle
{\let\thefootnote\relax\footnotetext{Accepted for publication in \emph{Philosophical Transactions of the Royal Society A}, theme issue `Safe, secure and robust AI for safety-critical systems'. This arXiv version is the authors' manuscript, typeset without the journal template. \copyright{} 2026 The Authors; distributed under the terms of the Creative Commons Attribution Licence (CC BY 4.0).}}

\begin{abstract}
Weak supervision enables machine learning models to learn from limited or noisy labels, but it introduces challenges in reliability and semantic clarity, particularly in multi-instance partial label learning (MI-PLL), where models must resolve both ambiguous supervision signals and uncertain instance-label mappings. This paper proposes a semantics for a neuro-symbolic framework that integrates inductive logic programming (ILP) to structure MI-PLL through relational constraints. In this formulation, ILP defines a hypothesis space over label transitions, formalizes the semantics of per-instance classifiers and provides a relational scaffold for reasoning about weak supervision. Two inductive tasks are studied in this framework: inferring the transition predicate from the observed and classifier predicates, and inferring instance-level classifier assignments from the observed and transition predicates. This formal semantics facilitates constraint specification, consistency checking and the diagnosis of semantic failure modes that bag-level accuracy alone may conceal.
\end{abstract}


\section{Introduction}
Weak supervision is an approach to machine learning for training models in settings where fully labelled data are limited. By reducing reliance on detailed labels, it enables models to learn from data that contain noise, ambiguity or missing labels \cite{brief_weak_supervision}. However, the benefits of weak supervision come with challenges: models trained under weak supervision can lack reliability and diagnosability, especially without a structured logical framework to guide learning. This limitation is particularly concerning in high-stakes applications where model accountability matters \cite{sheth2023neurosymbolic, garcez2020neurosymbolic}. Zhou \cite{brief_weak_supervision} categorizes weak supervision into three main types: incomplete supervision (only part of the data is labelled), inaccurate supervision (labels may contain errors) and inexact supervision (where labels are provided at a coarse-grained level). Within inexact supervision, multi-instance partial label learning (MI-PLL) is one such setting. MI-PLL combines multi-instance learning, where training samples consist of bags of instances, with partial label learning, where labels are ambiguous. In this setup, each training bag is associated with a candidate label set that includes at least one true label and potentially some false positives \cite{tang2024}.

Wang \textit{et~al.} \cite{wang2024learning} proved sample-complexity bounds for classifier learnability in MI-PLL. Their work explored cases where transition functions are either known or unknown, extending the known-transition analysis to multiple classifiers across diverse domains. The analysis combines classical statistical learning theory (Vapnik--Chervonenkis (VC) and Natarajan dimensions, and Rademacher complexity) with a weighted-model-count semantic loss for symbolic-label structure, yielding both theoretical bounds and an empirical evaluation. Building directly on that line, Tsamoura \textit{et~al.} \cite{tsamoura2025imbalances} extend the analysis to \emph{class-specific} risk in MI-PLL, showing that the symbolic component $\sigma$ can induce significant per-class learning imbalances even when the hidden gold-label distribution is uniform. Their class-specific risk bounds drop the $M$-unambiguity requirement of Wang \textit{et~al.}'s lemma~1 \cite{wang2024learning} and recover it in the $M$-unambiguous $M = 2$ case (Tsamoura \textit{et~al.}'s proposition~3.4), constituting a strict generalization of the latter (Wang \textit{et~al.}'s \cite{wang2024learning} other learnability results, including the sample-complexity theorems~1 and~5 and proposition~1, and the Rademacher-style top-$k$ surrogate theorem~2, are not subsumed). Together, these two works form the theoretical lineage on which the present framework builds: Wang \textit{et~al.} \cite{wang2024learning} supply the unambiguity conditions used to compare aggregation operators, and Tsamoura \textit{et~al.} \cite{tsamoura2025imbalances} supply the class-specific risk framework that provides formal language for interpreting the operator-dependent failure modes observed empirically.

This paper proposes a semantics for a neuro-symbolic framework that uses \MakeLowercase{Inductive Logic Programming} (ILP)~\cite{muggleton1994} to structure MI-PLL through relational constraints. The objectives are threefold: first, to logically characterize the hypothesis space for learning the transition function, examining how weak supervision operates given coarse labels and neural classifier outputs; second, to develop a logical framework that formalizes classifier semantics, defining both the targets and the functionalities of classifiers within weakly supervised settings; and third, to define a standard for assessing model behaviour against those relational constraints.

By structuring weak supervision within ILP, the noisy and ambiguous outputs of that supervision are systematically organized into relational constraints. This semantics is proposed in general terms, independent of any specific neuro-symbolic system: at the architectural level it is a `Neuro $\mid$ Symbolic' system in Kautz's \cite{kautz2022third} neuro-symbolic taxonomy, where the neural classifier maps non-symbolic image pixels to per-instance digit predictions and the CP/TP/OP machinery (the classifier predicate (CP), transition predicate (TP) and observed predicate (OP), formalized in \hyperref[sec:problem-structure-and-predicate-formulation]{\S2b}) then checks whether these predictions, the candidate transition rule and the bag-level observation are logically consistent. To test whether this semantics yields useful learning behaviour empirically, a differentiable pipeline in the style of the $\alpha$-ILP system is used as the experimental backend: the consistency check is made differentiable and gradients backpropagate to the classifier in joint training, which places the realization in Kautz's `$\text{Neuro}_{\text{Symbolic}}$' category (Kautz's category for differentiable logic, into which differentiable ILP falls). The semantics is the proposal; the differentiable realization is the experimental methodology through which it is tested.

\section{Logical foundations of weakly supervised induction}
This section develops the logical foundations of the framework, introducing symbolic notation and the bag-level supervision setting (\hyperref[sec:symbolic-semantics]{\S2a}), formalizing the MI-PLL problem with three auxiliary predicates CP, TP and OP (\hyperref[sec:problem-structure-and-predicate-formulation]{\S2b}) and describing ILP-based hypothesis induction in this setting (\hyperref[sec:ilp-based-hypothesis-induction]{\S2c}).

\subsection{Symbolic semantics of input data}
\label{sec:symbolic-semantics}
MNIST handwritten-digit images serve as a running example throughout the paper~\cite{lecun-gradientbased-learning-applied-1998, deng2012mnist}. For each digit \(j \in \{0, \ldots, 9\}\), \digitbox{j} denotes an MNIST image of that digit:

\[
\digitbox{j} \in \left\{
\mnistimg{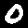},\,
\mnistimg{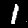},\,
\mnistimg{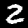},\,
\mnistimg{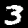},\,
\mnistimg{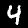},\,
\mnistimg{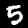},\,
\mnistimg{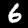},\,
\mnistimg{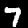},\,
\mnistimg{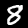},\,
\mnistimg{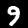}\right\}.
\]
Each \digitbox{j} has ground-truth label \(j\), visible by inspection but treated as latent by the model. When \digitbox{j} appears in two different bags, the occurrences are distinct exemplars of digit \(j\). Throughout the paper, \(j \in \{0, \ldots, 9\}\) ranges over digit values, while \(i \in \{1, \ldots, M\}\) indexes the position of an instance within a bag.

As the running example used in \hyperref[sec:problem-structure-and-predicate-formulation]{\S2b}, take the bag \((\mnistimg{2}, \mnistimg{3})\). A bag is written as a tuple of the constituent images, so \((\mnistimg{2}, \mnistimg{3})\) is the two-instance bag containing the images \(\mnistimg{2}\) and \(\mnistimg{3}\). Each image has an \emph{instance-level} latent label: \(\text{label}(\mnistimg{2}) = 2\) and \(\text{label}(\mnistimg{3}) = 3\). Under the aggregation operator \(\sigma = +\) (sum), the bag is supervised by a single \emph{bag-level} weak label \(s = \sigma(2, 3) = 5\), the only supervision the model receives. For compactness, the combined datum (the bag together with the observed weak label) is written as the \emph{sample form} \((\mnistimg{2}, \mnistimg{3}, 5)\): a tuple that lists the bag's \(M\) instance images first, followed by the bag-level label \(s\); this is the same training datum that the predicate atom \(\text{OP}((\mnistimg{2}, \mnistimg{3}), 5)\) records once predicates are introduced in \hyperref[sec:problem-structure-and-predicate-formulation]{\S2b}. In general, an MI-PLL training example has one latent instance-level label per image and one observed bag-level weak label per bag; the model must invert the aggregation across many bags to recover the instance labels (and, in some scenarios, the operator \(\sigma\)).

\subsection{Problem structure and predicate formulation}
\label{sec:problem-structure-and-predicate-formulation}

Formally, an MI-PLL instance is a tuple \((X, Y, S, M, \sigma)\) where \(X \subseteq \mathbb{R}^n\)~\cite{tang2024, brief_weak_supervision} is the instance space, \(Y\) is a discrete label space, \(S\) is a bag-level label space, \(M \geq 1\) is a bag size and \(\sigma : Y^M \to S\) is an aggregation operator. A training sample is a pair \((\mathbf{x}, s) \in X^M \times S\), where \(\mathbf{x} = (x_1, \ldots, x_M)\) is a bag of \(M\) instances with latent gold-label vector \(\mathbf{y} = (y_1, \ldots, y_M) \in Y^M\) (one label \(y_i\) per instance \(x_i\)), and \(s = \sigma(\mathbf{y})\) is the bag-level weak label obtained by applying \(\sigma\) to \(\mathbf{y}\). Only \((\mathbf{x}, s)\) is observed by the learner; \(\mathbf{y}\) is hidden. The learner accesses a finite training set \(\mathcal{T}_{\textsf{P}} = \{(\mathbf{x}^{(b)}, s^{(b)})\}_{b=1}^{N}\) of \(N\) such samples drawn independently and identically distributed (i.i.d.) from a distribution \(\mathcal{D}_{\textsf{P}}\) over \(X^M \times S\). The operators used in this setup (\(\sigma \in \{+, \times, \oplus\}\)) are permutation-invariant in \(\mathbf{y}\); the Boolean operator \(C\) introduced in \hyperref[sec:learning-scenarios]{\S}\ref{sec:learning-scenarios} (case~2 of scenario~2) is only partially permutation-invariant (\hyperref[lem:booleanC]{lemma}~\ref{lem:booleanC}) and is included to illustrate how the framework extends to operators with distinguished argument positions.

The three relationships in this setup, per-instance prediction, aggregation and bag-level observation, are abstracted as three auxiliary logical predicates (\emph{atom} is used throughout for a predicate applied to specific terms, e.g., \(\text{CP}(f, \mnistimg{2}, 2)\) is an atom of the CP predicate):

\begin{enumerate}
\item[---] \textbf{CP:} for a classifier \(f : X \to Y\), the atom \(\text{CP}(f, x_i, y_i) \equiv f(x_i) = y_i\) records that \(f\) assigns label \(y_i \in Y\) to instance \(x_i \in X\). For an MNIST input, the CP atom takes the form \(\text{CP}(f, \digitbox{j}, y)\) where \(y \in Y\) is the label assigned by \(f\); if \(f\) is correct, \(y\) equals the digit \(j\) shown by \(\digitbox{j}\). For example, \(\text{CP}(f, \mnistimg{2}, 2)\) means \(f\) correctly predicts label \(2\) on the image \(\mnistimg{2}\); \(\text{CP}(f, \mnistimg{2}, 4)\) means \(f\) predicts label \(4\) on the same image, which is incorrect since the digit shown is \(2\). CP is the predicate inferred in scenario~2 of \hyperref[sec:learning-scenarios]{\S}\ref{sec:learning-scenarios} (where \(\sigma\) is given and the per-instance assignment is learned).

\item [---]\textbf{TP:} for the aggregation operator \(\sigma : Y^M \to S\) of the MI-PLL setup, the atom \(\text{TP}_\sigma(\mathbf{y}, s) \equiv \sigma(\mathbf{y}) = s\) records that \(\sigma\) maps the instance label vector \(\mathbf{y} \in Y^M\) to the bag label \(s \in S\). For example, \(\text{TP}_+((2, 3), 5)\) holds because \(2 + 3 = 5\), whereas \(\text{TP}_+((2, 3), 6)\) does not hold (\(2 + 3 \neq 6\)). TP is the predicate inferred in scenario~1 (where \(\sigma\) is unknown and chosen from a candidate set \(\Sigma\)).

\item [---]\textbf{OP:} the atom \(\text{OP}(\mathbf{x}, s)\) records that the bag \(\mathbf{x} \in X^M\) is observed with weak label \(s \in S\); equivalently, \(\text{OP}(\mathbf{x}, s)\) holds for each training pair \((\mathbf{x}, s) \in \mathcal{T}_{\textsf{P}}\). For an MNIST bag, each \(x_i\) is a digit image; for example, \(\text{OP}((\mnistimg{2}, \mnistimg{3}), 5)\) is the OP atom for the running training sample \((\mnistimg{2}, \mnistimg{3}, 5)\), whereas \(\text{OP}((\mnistimg{2}, \mnistimg{3}), 7)\) would not hold for that sample since the observed weak label is \(5\), not \(7\). OP is the only directly observable predicate and anchors the supervision.
\end{enumerate}

\paragraph*{Notation conventions (summary)} Two conventions apply throughout the paper: (i) a \emph{sample form} \((x_1, \ldots, x_M, s)\) for enumerating training data, equivalent to the atom \(\text{OP}((x_1, \ldots, x_M), s)\) and (ii) an \emph{operator-subscript} convention where the symbols \(\text{TP}\), \(\mathbb{I}\) and \(\mathbb{H}\) take a generic subscript (\(\text{TP}_\sigma, \mathbb{I}_\sigma\text{ and } \mathbb{H}_\sigma\)) when the operator is hypothesized or being searched over (scenario~1 of \hyperref[sec:learning-scenarios]{\S}\ref{sec:learning-scenarios}), and a specific subscript (\(\text{TP}_+, \text{TP}_\times, \text{TP}_\oplus, \text{TP}_C\), and likewise for \(\mathbb{I}\) and \(\mathbb{H}\)) when the operator is fixed and given (scenario~2). Full statements of both conventions are given in \hyperref[app:proofs]{appendix B}.

Continuing the running example from \hyperref[sec:symbolic-semantics]{\S2a}, the training datum \((\mnistimg{2}, \mnistimg{3}, 5)\) is recorded by the OP atom \(\text{OP}((\mnistimg{2}, \mnistimg{3}), 5)\). Under \(\sigma = +\) with \(s = 5\), the TP atom \(\text{TP}_\sigma((y_1, y_2), 5)\) is satisfied by label pairs including \((5, 0)\), \((1, 4)\) and \((2, 3)\). Suppose the classifier \(f\) outputs the correct labels: \(f(\mnistimg{2}) = 2\) and \(f(\mnistimg{3}) = 3\), recorded by the CP atoms \(\text{CP}(f, \mnistimg{2}, 2)\) and \(\text{CP}(f, \mnistimg{3}, 3)\). Plugging the predicted labels into \(\sigma\) gives \(\sigma(2, 3) = 5 = s\), so \(\text{TP}_\sigma((2, 3), 5)\) holds and the three atoms \((\text{CP}, \text{TP}_\sigma\text{ and }\text{OP})\) are jointly satisfied. If instead \(f\) misclassifies (\(f(\mnistimg{2}) = 4\), \(f(\mnistimg{3}) = 2\)), the CP atoms become \(\text{CP}(f, \mnistimg{2}, 4)\) and \(\text{CP}(f, \mnistimg{3}, 2)\), and \(\sigma(4, 2) = 6 \neq 5\), so \(\text{TP}_\sigma((4, 2), 5)\) fails and the three atoms are inconsistent. This first example pairs every atom with both sample and predicate forms; subsequent examples enumerate training data in sample form and assemble predicate atoms inside constraint structures and consistency checks without restating the mapping. The pair of cases shown here illustrates the consistency check on a single bag; in practice, learning involves many bags and a hypothesis \(H = (f, \sigma)\) drawn from a search space, and the goal is to find an \(H\) under which the three atoms are jointly satisfied on as many bags as possible. \hyperref[sec:ilp-based-hypothesis-induction]{Section 2c} describes how ILP structures this search.

\subsection{Inductive logic programming-based hypothesis induction}
\label{sec:ilp-based-hypothesis-induction}
ILP is a relational learning framework that induces logic programs from data, using background knowledge to constrain the hypothesis space \cite{muggleton1994, cropper2022inductive}. Formally, given a background theory \( B \), a set of positive examples \( E^+ \), and a set of negative examples \( E^- \), the goal is to find a hypothesis \( H \), a set of clauses, such that \( B \cup H \models e \) for all \( e \in E^+ \), and \( B \cup H \not\models e \) for all \( e \in E^- \). The hypothesis \( H \) is drawn from a restricted language defined by a syntactic bias (e.g. mode declarations and maximum clause length), ensuring tractability and domain relevance.

MI-PLL is cast as an ILP problem with a structured hypothesis. A hypothesis is a pair \(H = (f, \sigma)\), where \(f : X \to Y\) is a classifier and \(\sigma\) is an aggregation operator drawn from the candidate set \(\Sigma\) (e.g. \(\Sigma = \{+, \times, \oplus, C\}\)). The background knowledge \(B\) encodes \(\Sigma\) together with each operator's structural constraints (permutation invariance for \(+, \times, \oplus\); partial permutation invariance for \(C\); \hyperref[lem:booleanC]{lemma}~\ref{lem:booleanC}). Each training bag \((\mathbf{x}, s) \in \mathcal{T}_{\textsf{P}}\) supplies the OP atom \(\text{OP}(\mathbf{x}, s)\) as supervision. Given a hypothesis \(H = (f, \sigma)\), each bag is classified as a positive example (\(\mathbf{x} \in E^+\)) if the CP atoms induced by \(f\) jointly satisfy \(\text{TP}_\sigma((f(x_1), \ldots, f(x_M)), s)\), and as a negative example (\(\mathbf{x} \in E^-\)) otherwise. The partition is therefore hypothesis-dependent; the formal definition is given in \hyperref[sec:learning-scenarios]{\S}\ref{sec:learning-scenarios}.

Learning in this setting is both partially observable and abductive. Since instance labels \(y_i\) are hidden, the learner must abduce latent CP and TP assignments such that these jointly entail the observed OP atoms; the resulting bag-by-bag conjunction is called an \emph{intermediate interpretation} \(\mathbb{I}_\sigma\), parametrized by the operator under hypothesis. The ILP task is to search the hypothesis space \(\mathbb{H}\) of pairs \(H = (f, \sigma)\) for one whose intermediate interpretation \(\mathbb{I}_\sigma\) satisfies the supervision \(B \cup H \models \text{OP}(\mathbf{x}, s)\) on as many training bags as possible (\hyperref[sec:learning-scenarios]{\S3} replaces this entailment reading with a hypothesis-dependent partition, since the OP atoms are observations rather than derivable facts).

To illustrate this process, consider four training examples in sample form: \((\mnistimg{2}, \mnistimg{3}, 7)\), \((\mnistimg{1}, \mnistimg{4}, 5)\), \((\mnistimg{5}, \mnistimg{2}, 9)\) and \((\mnistimg{3}, \mnistimg{6}, 8)\). Assuming the aggregation operator \(\sigma = +\), each pair of latent labels is hypothesized to sum to the corresponding weak label. Suppose \(f\) is a perfect classifier, so \(\text{CP}(f, \digitbox{j}, j)\) holds for every digit \(j\). Bag 2 is then consistent under \(\sigma = +\) since \(\text{TP}_\sigma((1, 4), 5)\) holds. Bags 1, 3 and 4 fail: \(2 + 3 = 5 \neq 7\), \(5 + 2 = 7 \neq 9\) and \(3 + 6 = 9 \neq 8\). Thus, only Bag 2 is consistent with the hypothesized addition operator under these CP predictions.

More generally, the internal structure induced by this reasoning yields one positive clause and three violations. This symbolic structure is represented as a conjunction of classifier outputs and aggregation constraints under \(\sigma = +\), denoted by:
\[
\mathbb{I}_\sigma = \left(
\begin{aligned}
&\text{CP}(f, \mnistimg{2}, 2) \land \text{CP}(f, \mnistimg{3}, 3) \land \neg\,\text{TP}_\sigma((2, 3),\, 7) \\
\land\;& \text{CP}(f, \mnistimg{1}, 1) \land \text{CP}(f, \mnistimg{4}, 4) \land \text{TP}_\sigma((1, 4),\, 5) \\
\land\;& \text{CP}(f, \mnistimg{5}, 5) \land \text{CP}(f, \mnistimg{2}, 2) \land \neg\,\text{TP}_\sigma((5, 2),\, 9) \\
\land\;& \text{CP}(f, \mnistimg{3}, 3) \land \text{CP}(f, \mnistimg{6}, 6) \land \neg\,\text{TP}_\sigma((3, 6),\, 8)
\end{aligned}
\right).
\]

This structure captures both satisfied and violated constraints under the hypothesized aggregation rule \(\sigma = +\). Such discrepancies help the ILP learner to assess the plausibility of \(\sigma\) as the governing rule. The resulting internal structure, denoted \(\mathbb{I}_\sigma\), serves as symbolic evidence that links classifier outputs (CP), instance-level label assignments and observed bag-level labels (OP) through an induced TP.
\paragraph*{Constraint instances and conjunction} Each line contributing to $\mathbb{I}_\sigma$ corresponds to a distinct constraint instance induced by a different bag. The conjunction notation denotes the collection of all such constraints across bags, rather than a single logical
clause. Predicate atoms such as $\mathrm{CP}(f,x,y)$ may therefore appear
multiple times without redundancy, since each occurrence refers to a different
data instance. Logical conjunction expresses the requirement that all
bag-level constraints be considered simultaneously during hypothesis
evaluation.

Inducing and evaluating \(\mathbb{I}_\sigma\) within the ILP framework generates symbolic hypotheses that link latent label assignments to the observed weak supervision. At the semantic level, ILP can be viewed as playing two complementary roles: a generative one, postulating plausible latent structures consistent with the weak supervision; a deductive one, filtering those structures by logical entailment of the observed OP atoms. The implementation realizes this view through differentiable soft marginalization over the TP and a bag-level cross-entropy loss against the observed weak labels, rather than through explicit symbolic abductive search; the predicate-level decomposition still describes what the joint training approximates. \hyperref[sec:learning-scenarios]{Section }{\ref{sec:learning-scenarios}} specializes the \((f, \sigma)\) decomposition into two scenarios: scenario~1 fixes \(f\) and infers \(\sigma\); scenario~2 fixes \(\sigma\) and infers \(f\).

\section{Learning scenarios in multi-instance partial label
learning }\label{sec:learning-scenarios}
MI-PLL is formalized as an ILP task by encoding the hidden relationships between bag-level supervision and latent instance-level labels as logical hypotheses over the three predicates \textbf{CP}, \textbf{TP} and \textbf{OP}, forming a structured hypothesis space.

Two core learning scenarios are considered, each posing a different inductive task based on partial observability:

\vspace{0.5em}
\noindent\textbf{Scenario 1: inferring \(\text{TP}\) from \(\text{OP}\) and \(\text{CP.}\)}
This scenario captures the case where the weak label is known (OP), and instance-level predictions (CP) are assumed to be available. The goal is to discover a transition rule, captured by TP, that explains how instance labels collectively yield the bag-level label.
\begin{equation}
H_{\text{TP}} \equiv \forall (\mathbf{x}, s) \in \mathcal{T}_{\textsf{P}}, \;
\forall \mathbf{y} \in Y^M, \;
\left( \text{OP}(\mathbf{x}, s) \land \bigwedge_{i=1}^{M} \text{CP}(f, x_i, y_i) \right)
\Rightarrow \text{TP}_\sigma((y_1, \ldots, y_M), s),
\label{eq:htp}
\end{equation}
where the label vector \(\mathbf{y}\) is universally quantified; since
\(\text{CP}(f, x_i, \cdot)\) holds for exactly one value (\(y_i = f(x_i)\)),
the formula is equivalent to requiring
\(\text{TP}_\sigma((f(x_1), \ldots, f(x_M)), s)\) on every training bag,
which is also the content of \linkref[equation]{eq:hcp} below; the two
scenarios share this consistency condition and differ in which component
(\(\sigma\) or \(f\)) is searched.
\vspace{0.5em}
\noindent\textbf{Scenario 2: inferring \(\text{CP}\) from \(\text{OP}\) and \(\text{TP.}\)}
Here, the transition rule is treated as given, and the goal is to recover a classifier \(f\) whose per-instance predictions \(y_i := f(x_i)\) are jointly consistent with the bag-level label under the known TP. Writing \(\mathbf{f}(\mathbf{x}) := (f(x_1), \ldots, f(x_M))\), the hypothesis is given by:
\begin{equation}
H_{\text{CP}} \equiv \forall (\mathbf{x}, s) \in \mathcal{T}_{\textsf{P}}, \;
\text{OP}(\mathbf{x}, s) \;\Rightarrow\; \text{TP}_\sigma(\mathbf{f}(\mathbf{x}), s).
\label{eq:hcp}
\end{equation}
Each hypothesis \(H = (f, \sigma)\) represents a structured explanation over latent components and is refined using sets of positive and negative examples (in both sets the latent labels are fixed by the classifier, \(y_i := f(x_i)\), which closes the formulas):
\[
E^+ = \Big\{ (\mathbf{x}, s) \in \mathcal{T}_{\textsf{P}} \;\Big|\; \text{OP}(\mathbf{x}, s) \land \bigwedge_{i=1}^{M} \text{CP}(f, x_i, y_i) \land \text{TP}_\sigma((y_1, \ldots, y_M), s) \Big\}
\]
\[
\text{and }E^- = \Big\{ (\mathbf{x}, s) \in \mathcal{T}_{\textsf{P}} \;\Big|\; \text{OP}(\mathbf{x}, s) \land \bigwedge_{i=1}^{M} \text{CP}(f, x_i, y_i) \land \neg\,\text{TP}_\sigma((y_1, \ldots, y_M), s) \Big\}.
\]

Hypotheses are evaluated against these examples, with the hypothesis
language restricted by the background knowledge \(B\), optimizing the
partition gap:
\[
\max_{H} \left( |E^+(H)| - |E^-(H)| \right).
\]

This objective rewards hypotheses whose induced predictions cover as many
observed OP atoms as possible while violating as few as possible; the
entailment relation \(B \cup H \models e\) of classical ILP is replaced
here by the hypothesis-dependent partition, since the OP atoms are
observations rather than derivable facts.

\paragraph*{Abductive partition} The sets \(E^+\) and \(E^-\) are
\emph{hypothesis-dependent}: each bag's membership is determined by the CP
predictions and TP rule contained in the current hypothesis \(H\). Unlike
classical ILP, where \(E^+\) and \(E^-\) are static problem inputs, here, the
partition shifts as the abduced latent assignments \(y_i\) and candidate
operator \(\sigma\) evolve during search. The objective above is therefore a
joint optimization over \(H\) and the induced partition that \(H\) defines.

\paragraph*{Reduction under the (CP and TP) decomposition} Parametrizing \(H\)
as a pair \((f, \sigma)\) makes the objective equivalent to bag-level
classification accuracy on the weak labels: for every bag, \(\sigma(f(\mathbf{x}))=s\)
places it in \(E^+\), otherwise in \(E^-\). Since \(|E^+| + |E^-|\) is the
total number of supervised bags \(N\) (bags being pairwise distinct;
\hyperref[app:proofs]{appendix~B}), maximizing \(|E^+| - |E^-| = 2|E^+| - N\)
is monotone in \(|E^+|\) and hence in bag accuracy. What this paper adds
is therefore not the objective form but the \emph{structured
hypothesis space}: \(H\) decomposes into a neural classifier \(f\) (CP) and a
symbolic operator \(\sigma\) (TP) drawn from a finite candidate set, with the
search restricted by background knowledge \(B\), yielding a hypothesis whose
internal structure is logically decomposed rather than monolithic.

\subsection{Scenario 1: inferring the transition predicate}

In this scenario, the goal is to infer the \(\text{TP}\) from the observed bag-level supervision \(\text{OP}\) and the instance-level classifier predictions \(\text{CP}\) as described in \linkref[equation]{eq:htp}. The true aggregation operator \(\sigma\) that maps instance labels to a bag label is not known in advance. Rather than assuming a fixed aggregation rule, a finite set of candidate operators \(\sigma \in \{+, \times, \oplus, C\}\) is considered, with the aim of identifying the operator that best explains the relationship between predicted instance labels and observed bag-level outputs. This is achieved by systematically evaluating each candidate \(\text{TP}_\sigma\) against the observed examples and selecting the most consistent TP.

\textbf{Case 1 (\(\sigma = \times\))}: consider four training examples in sample form: \((\mnistimg{2}, \mnistimg{3}, 6)\), \((\mnistimg{4}, \mnistimg{3}, 12)\), \((\mnistimg{5}, \mnistimg{2}, 11)\) and \((\mnistimg{4}, \mnistimg{2}, 7)\). Let the classifier \(f\) assign predictions: \(\text{CP}(f, \mnistimg{2}, 2)\), \(\text{CP}(f, \mnistimg{3}, 3)\), \(\text{CP}(f, \mnistimg{4}, 4)\) and \(\text{CP}(f, \mnistimg{5}, 5)\). Under the hypothesis that \(\sigma = \times\), the candidate predicate \(\text{TP}_\sigma\) is evaluated. Bags 1 and 2 are consistent: \(2 \times 3 = 6\), \(4 \times 3 = 12\). Bags 3 and 4, however, are inconsistent: \(5 \times 2 = 10 \neq 11\), \(4 \times 2 = 8 \neq 7\). Hence, \(\sigma = \times\) explains two of four examples.

\[
\mathbb{I}_\sigma = \left(
\begin{aligned}
& \text{CP}(f, \mnistimg{2}, 2) \land \text{CP}(f, \mnistimg{3}, 3) \land \text{TP}_\sigma((2, 3), 6) \\
\land\;& \text{CP}(f, \mnistimg{4}, 4) \land \text{CP}(f, \mnistimg{3}, 3) \land \text{TP}_\sigma((4, 3), 12) \\
\land\;& \text{CP}(f, \mnistimg{5}, 5) \land \text{CP}(f, \mnistimg{2}, 2) \land \neg\,\text{TP}_\sigma((5, 2), 11) \\
\land\;& \text{CP}(f, \mnistimg{4}, 4) \land \text{CP}(f, \mnistimg{2}, 2) \land \neg\,\text{TP}_\sigma((4, 2), 7)
\end{aligned}
\right).
\]

\textbf{Case 2 (\(\sigma = +\))}: consider four training examples in sample form: \((\mnistimg{4}, \mnistimg{7}, 11)\), \((\mnistimg{3}, \mnistimg{6}, 9)\), \((\mnistimg{2}, \mnistimg{9}, 11)\) and \((\mnistimg{4}, \mnistimg{1}, 6)\). The classifier's predictions are given as: \(\text{CP}(f, \mnistimg{4}, 4)\), \(\text{CP}(f, \mnistimg{7}, 7)\), \(\text{CP}(f, \mnistimg{3}, 3)\), \(\text{CP}(f, \mnistimg{6}, 6)\), \(\text{CP}(f, \mnistimg{2}, 2)\), \(\text{CP}(f, \mnistimg{9}, 9)\) and \(\text{CP}(f, \mnistimg{1}, 1)\). Assuming \(\sigma = +\), bags 1--3 are satisfied: \(4 + 7 = 11\), \(3 + 6 = 9\), \(2 + 9 = 11\); bag 4 is violated: \(4 + 1 = 5 \neq 6\). The resulting structure is:
\[
\mathbb{I}_\sigma = \left(
\begin{aligned}
& \text{CP}(f, \mnistimg{4}, 4) \land \text{CP}(f, \mnistimg{7}, 7) \land \text{TP}_\sigma((4, 7), 11) \\
\land\;& \text{CP}(f, \mnistimg{3}, 3) \land \text{CP}(f, \mnistimg{6}, 6) \land \text{TP}_\sigma((3, 6), 9) \\
\land\;& \text{CP}(f, \mnistimg{2}, 2) \land \text{CP}(f, \mnistimg{9}, 9) \land \text{TP}_\sigma((2, 9), 11) \\
\land\;& \text{CP}(f, \mnistimg{4}, 4) \land \text{CP}(f, \mnistimg{1}, 1) \land \neg\,\text{TP}_\sigma((4, 1), 6)
\end{aligned}
\right).
\]

In both illustrative cases, the underlying aggregation \(\sigma : Y^M \to S\)
is a function, so \(\text{TP}\) is functional in \(\mathbf{y}\); however, \(\sigma\)
may be non-injective, meaning that for a given bag label \(s\), the pre-image
\(\sigma^{-1}(s) = \{\mathbf{y} \in Y^M : \sigma(\mathbf{y}) = s\}\) can contain
multiple instance-label tuples. This non-injectivity (not the predicate
form) is what permits multiple valid instance-to-bag mappings. For each candidate \(\sigma \in \{+, \times, \oplus, C\}\), a TP \(\text{TP}_\sigma\) is instantiated and evaluated using the observed supervision (\(\text{OP}\)) and classifier predictions (\(\text{CP}\)). This search process is formalized by selecting a candidate that maximizes consistency on the positive examples: choose \(\sigma \in \{+, \times, \oplus, C\}\) to maximize
\[
\bigl|\bigl\{(\mathbf{x}, s) \in \mathcal{T}_{\textsf{P}} \;:\; \text{TP}_\sigma((y_1, \ldots, y_M), s) \text{ holds with } y_i = f(x_i) \bigr\}\bigr|,
\]
where for each bag the latent labels are fixed by the CP atoms (\(y_i = f(x_i)\)).
Negative examples \((\mathbf{x}, s) \in E^-\), when present, contribute through
the dual condition \(\neg\,\text{TP}_\sigma((y_1, \ldots, y_M), s)\), so the
joint ILP objective is the partition gap \(|E^+(\sigma)| - |E^-(\sigma)|\)
of \hyperref[app:proofs]{appendix B}.
The ILP engine then selects the most consistent and explanatory hypothesis \(\mathbb{H}_\sigma\), an operator-level hypothesis that explains bag-level labels via logical aggregation over instance-level predictions. This procedure systematically refines and validates symbolic hypotheses in weakly supervised settings.

\paragraph*{Operator-by-operator learnability predictions} The MI-PLL learnability framework of Wang \textit{et~al.}\ \cite{wang2024learning} and the class-specific refinement by Tsamoura \textit{et~al.}~\cite{tsamoura2025imbalances} predict different difficulty regimes for each of the three operators through two structural conditions on \(\sigma\) from the former framework: \emph{\(M\)-unambiguity} (distinct diagonal label vectors yield distinct bag labels) and the logically independent \emph{\(1\)-unambiguity} (some position \(i\) such that flipping \(y_i\) always changes \(\sigma(\mathbf{y})\)). The three operators fall into three cells of this profile: \(\sigma = +\) is both \(1\)- and \(M\)-unambiguous (fastest sample-complexity rate of Wang \textit{et~al.}\ proposition~1 \cite{wang2024learning}); \(\sigma = \times\) is \(M\)-unambiguous but not \(1\)-unambiguous (the slower \(M\)-th-root rate of Wang \textit{et~al.}\ theorem~1 \cite{wang2024learning}, and the formal condition associated with the multiplicative zero-attractor observed); \(\sigma = \oplus\) on \(M = 2\) is \(1\)-unambiguous but not \(M\)-unambiguous (no aggregate-risk bound applies; the configuration is consistent with the failure mode Wang \textit{et~al.}'s lemma~1 establishes existentially). Wang \textit{et~al.}'s \cite{wang2024learning} bounds are on sample complexity, not optimization dynamics; the empirical patterns are \emph{consistent with} the predicted regimes rather than direct tests of those regimes. Full restatements, proofs and the detailed framework mapping are deferred to \hyperref[app:wang]{appendix C} (Wang) and \hyperref[app:tsamoura]{appendix D} (Tsamoura's class-specific extension via propositions~3.1 and~3.4).

\subsection{Scenario 2: inferring the classifier predicate}
\label{sec:scenario2}
In this scenario, the goal is to infer the classifier \(f : X \to Y\), recorded by the CP atoms \(\text{CP}(f, x_i, y_i)\) for each instance \(x_i\) in each training bag. The aggregation operator \(\sigma\) is assumed known and given by the fixed TP \(\text{TP}_\sigma\). Given observed OP atoms \(\text{OP}((x_1, x_2, \ldots), s)\) and a fixed \(\text{TP}_\sigma\), the task is to recover \(f\) so that the induced CP atoms are jointly consistent with the observed bag-level outcomes and the known transition rule.

\textbf{Case 1 (\(\sigma = \oplus\), XOR aggregation)}: suppose the aggregation operator is known to be bitwise XOR, so the TP satisfies \(\text{TP}_\oplus((y_1, y_2), s) \equiv (y_1 \oplus y_2 = s)\). In this setting, the bag-level label is obtained by applying the XOR operation to the digit predictions of two MNIST images within each bag. Given the following training examples in sample form, \((\mnistimg{3}, \mnistimg{5}, 6)\), \((\mnistimg{2}, \mnistimg{7}, 5)\), \((\mnistimg{4}, \mnistimg{1}, 5)\) and \((\mnistimg{6}, \mnistimg{7}, 1)\), the digit label for each instance is to be inferred such that these relationships hold. This leads to the following internal constraint structure:

\[
\mathbb{I}_\oplus = \left(
\begin{aligned}
&\text{CP}(f, \mnistimg{3}, y_1) \land \text{CP}(f, \mnistimg{5}, y_2) \land \text{TP}_\oplus((y_1, y_2), 6) \\
\land\;&\text{CP}(f, \mnistimg{2}, y_3) \land \text{CP}(f, \mnistimg{7}, y_4) \land \text{TP}_\oplus((y_3, y_4), 5) \\
\land\;&\text{CP}(f, \mnistimg{4}, y_5) \land \text{CP}(f, \mnistimg{1}, y_6) \land \text{TP}_\oplus((y_5, y_6), 5) \\
\land\;&\text{CP}(f, \mnistimg{6}, y_7) \land \text{CP}(f, \mnistimg{7}, y_4) \land \text{TP}_\oplus((y_7, y_4), 1)
\end{aligned}
\right).
\]

This constraint structure is now examined under the XOR operation. The first bag label \(6\) is explained by the pair \(3 \oplus 5 = 6\), the second by \(2 \oplus 7 = 5\), the third by \(4 \oplus 1 = 5\) and the fourth by \(6 \oplus 7 = 1\). These assignments jointly satisfy all bag-level XOR constraints, resulting in the classifier predictions: \(\text{CP}(f, \mnistimg{3}, 3)\), \(\text{CP}(f, \mnistimg{5}, 5)\), \(\text{CP}(f, \mnistimg{2}, 2)\), \(\text{CP}(f, \mnistimg{7}, 7)\), \(\text{CP}(f, \mnistimg{4}, 4)\), \(\text{CP}(f, \mnistimg{1}, 1)\) and \(\text{CP}(f, \mnistimg{6}, 6)\). Hence, the consistent classifier hypothesis under XOR aggregation is given by \(\mathbb{H}_{\oplus} \equiv \forall (\mathbf{x}, s) \in \mathcal{T}_{\textsf{P}}, \; \text{OP}(\mathbf{x}, s) \Rightarrow \text{TP}_\oplus(\mathbf{f}(\mathbf{x}), s)\), the form of \linkref[equation]{eq:hcp} specialized to \(\sigma = \oplus\) (writing \(\mathbf{f}(\mathbf{x}) = (f(x_1), \ldots, f(x_M))\)), meaning that the per-instance predictions \(f(x_i)\) are jointly consistent with the observed bag label under XOR.

The classifier predictions listed above are the witness for this hypothesis: under them every bag's XOR equals its observed label, so \(\text{TP}_\oplus(\mathbf{f}(\mathbf{x}), s)\) holds for all four bags.

\textbf{Case 2 (\(\sigma = C\), Boolean formula aggregation)}: now consider a scenario where the TP is governed by a Boolean formula \(\sigma_C(y_1, y_2, y_3) := (y_1 \land y_2) \lor (y_1 \land y_3)\), so \(\text{TP}_C((y_1, y_2, y_3), s) \equiv \sigma_C(y_1, y_2, y_3) = s\), with instance-level values restricted to \(\{0, 1\}\). Note that \(\text{TP}_C\) is symmetric in \(y_2\) and \(y_3\) but distinguishes \(y_1\); it is therefore only \emph{partially} permutation-invariant. This case is retained as a theoretical illustration of how hypothesis complexity grows when symbolic structure is non-uniform across instance positions. The experimental evaluation in \hyperref[sec:experiments-and-evaluation]{\S}\ref{sec:experiments-and-evaluation} restricts attention to fully permutation-invariant operators (\(+, \times, \oplus\)); operators with distinguished argument positions (such as \(\text{TP}_C\), subtraction and division) require an order-aware TP formulation and are left for future work. Given the following training examples in sample form, \((\mnistimg{1}, \mnistimg{3}, \mnistimg{2}, 1)\), \((\mnistimg{2}, \mnistimg{3}, \mnistimg{4}, 0)\), \((\mnistimg{1}, \mnistimg{4}, \mnistimg{3}, 1)\) and \((\mnistimg{2}, \mnistimg{1}, \mnistimg{4}, 0)\), the aim is to infer binary predictions per instance that collectively satisfy the Boolean structure. This gives rise to the constraint formula:
\[
\mathbb{I}_{C} = \left(
\begin{aligned}
&\text{CP}(f, \mnistimg{1}, y_1) \land
\text{CP}(f, \mnistimg{3}, y_2) \\
\land\; &\text{CP}(f, \mnistimg{2}, y_3) \land
\text{CP}(f, \mnistimg{4}, y_4) \\
\land\; &\text{TP}_C((y_1, y_2, y_3), 1) \land \text{TP}_C((y_3, y_2, y_4), 0) \\
\land\; &\text{TP}_C((y_1, y_4, y_2), 1) \land
\text{TP}_C((y_3, y_1, y_4), 0)
\end{aligned}
\right).
\]

One solution to this constraint structure assigns \(\text{CP}(f, \mnistimg{1}, 1)\), \(\text{CP}(f, \mnistimg{2}, 0)\), \(\text{CP}(f, \mnistimg{3}, 1)\) and \(\text{CP}(f, \mnistimg{4}, 0)\). Under this assignment, the Boolean formula \((y_1 \land y_2) \lor (y_1 \land y_3)\) evaluates to \(1\), \(0\), \(1\) and \(0\) for the four bags, respectively: specifically, \((1 \land 1) \lor (1 \land 0) = 1\), \((0 \land 1) \lor (0 \land 0) = 0\), \((1 \land 0) \lor (1 \land 1) = 1\) and \((0 \land 1) \lor (0 \land 0) = 0\). These match all observed bag-level labels.

This yields the classifier hypothesis \(\mathbb{H}_C \equiv \forall (\mathbf{x}, s) \in \mathcal{T}_{\textsf{P}}, \; \text{OP}(\mathbf{x}, s) \Rightarrow \text{TP}_C(\mathbf{f}(\mathbf{x}), s)\), again the form of \linkref[equation]{eq:hcp} specialized to \(\sigma = C\).

These case analyses illustrate how the task of inferring \(\text{CP}\) is shaped by the structural complexity of the TP \(\text{TP}_\sigma\). While the hypothesis for addition requires relatively straightforward additive consistency, those for multiplication, XOR (\(\mathbb{H}_{\oplus}\)) and Boolean aggregation (\(\mathbb{H}_C\)) demand increasingly complex symbolic reasoning. As the complexity of the aggregation rule increases, the ILP engine must operate over tighter and more expressive hypothesis spaces to maintain logical consistency between instance-level assignments and observed bag-level outputs.

\section{Experiments and evaluation}
\label{sec:experiments-and-evaluation}

To operationalize the proposed OP--CP--TP semantics, the experiments use a
differentiable pipeline in the style of $\alpha$-ILP, the differentiable ILP
framework that integrates neural predicates with symbolic hypothesis
scoring via differentiable logical reasoning~\cite{shindo2023alpha}. The
realization here follows that design rather than invoking the full
clause-induction machinery: neural predicates instantiate the CP,
differentiable marginalization through the candidate operators scores TP
hypotheses, and the OP facts enter as soft constraints
(\hyperref[alg:unified]{algorithm}~\ref{alg:unified}).

In the experiments, this pipeline serves as an execution substrate for testing
the proposed semantics rather than as the contribution. Neural perception is used to instantiate the CP
from raw images, while symbolic reasoning is used to represent and select
TPs from a predefined operator space. Weak supervision
enters exclusively through OPs, which are treated as
logical constraints rather than direct training targets. The
pipeline enforces consistency among these components by favouring hypotheses
whose induced CP and TP assignments jointly entail the observed OP facts.

The semantic claims do not rely on any architectural feature
specific to this realization. Any neurosymbolic system capable of (i) explicitly
representing CP, TP and OP as distinct predicates, (ii) reasoning over
symbolic hypothesis spaces, and (iii) enforcing logical consistency under weak
supervision could, in principle, instantiate the same semantics. The
differentiable pipeline provides a concrete realization that allows these
dependencies to be empirically evaluated.

The experimental design intentionally focuses on simple MI-PLL tasks in order
to isolate semantic dependencies and make failure modes observable, rather than
to maximize raw predictive performance.

\subsection{Experimental setup}
The framework is evaluated on MI-PLL datasets constructed from MNIST digit bags.
Each bag contains exactly two digit images with hidden instance labels. The only
supervision provided during training is a weak bag-level label, obtained by
applying a symbolic aggregation operator to the latent instance labels. The
experiments reported below cover the operator set,
\[
\sigma \in \{+, \times, \oplus\},
\]
chosen to span the Wang \textit{et~al.}\ \cite{wang2024learning} unambiguity profile
discussed in \hyperref[app:wang:unambig]{appendix C.2}: addition satisfies both
\(1\)- and \(M\)-unambiguity, multiplication satisfies \(M\)- but not
\(1\)-unambiguity and XOR satisfies \(1\)- but not \(M\)-unambiguity.
The choice also probes the class-specific imbalance regime
characterized by Tsamoura \textit{et~al.}~\cite{tsamoura2025imbalances}:
multiplication on the digit range has an asymmetric pre-image
distribution (the bag label \(s = 0\) absorbs 19 of the 100 ordered
pairs) while addition and XOR have flatter pre-image profiles, so
operator-by-operator differences in OP-noise tolerance and per-class
collapse become legible against the Tsamoura \textit{et~al.}\ framework
(\hyperref[app:tsamoura]{appendix D}). The evaluation
restricts to fully permutation-invariant operators here; order-sensitive
operators (subtraction and division) and the Boolean formula
\(\mathrm{TP}_C\) used illustratively in~\hyperref[sec:scenario2]{\S3b} require an
order-aware TP and are left for future work.

In the implementation, CP is realized as a convolutional neural network, TP is
drawn from a discrete hypothesis space of symbolic operators and the pipeline
mediates the CP--TP interaction by enforcing the CP/TP/OP consistency constraint of \hyperref[sec:problem-structure-and-predicate-formulation]{\S2b} and selecting
consistent hypotheses. All reported results are averaged over multiple random
seeds (20 runs for experiments~1 and~2, and 10 runs for the OP-noise study),
reported as mean $\pm$ standard deviation.

\paragraph*{Training details}
Each run uses \(N_\text{train} = 2000\) training bags and \(N_\text{val} = 500\)
validation bags drawn from MNIST. Supervision for each training bag is a
candidate set of size \(3\) containing the true weak label \(s_b\) and two
distractor labels drawn from the feasible bag-label range of \(\sigma\);
the OP distribution \(p_{\text{OP}}\) of
\hyperref[alg:unified]{algorithm}~\ref{alg:unified} is uniform over this
set. This realizes the candidate-set form of MI-PLL supervision
(\S1) with the aggregated label of
\hyperref[sec:problem-structure-and-predicate-formulation]{\S2b} as the
guaranteed true member. The weak label itself is left uncorrupted in
experiments~1 and~2; only the OP-noise study replaces \(s_b\) at the
stated corruption rates, and all validation metrics are computed against
clean weak labels throughout. The CP convolutional network is trained with Adam at
a learning rate of \(10^{-3}\) and batch size \(64\). The epoch budget is \(15\) for the scenario-2 driver
(\hyperref[tab:unified_ablation]{tables}~\ref{tab:unified_ablation} and~\ref{tab:exp2_cp}), \(10\)
CP-warm followed by \(15\) TP-discovery epochs for the scenario-1
driver (\hyperref[tab:exp1_tp]{table}~\ref{tab:exp1_tp}) and \(8\) CP-warm followed by \(15\)
TP-discovery epochs for the OP-noise study
(\hyperref[tab:op_noise]{table}~\ref{tab:op_noise}). In the scenario-1
drivers, the CP-warm phase trains the classifier under the true
operator (the scenario-2 configuration); TP discovery then freezes the
classifier and trains only the operator weights
(\hyperref[sec:experiments-and-evaluation]{\S4c}). The \hyperref[alg:unified]{algorithm}~\ref{alg:unified} four-term
loss is applied with weights
\((\lambda_{\text{bag}}, \lambda_{\text{con}}, \lambda_{\text{cp}},
\lambda_{\text{ctp}}) = (1.0,\ 1.0,\ 0.05,\ 0.5)\) and the CP--TP
distance realized as squared \(\ell_2\) between the CP-derived and
TP-derived bag distributions. Auxiliary entropy and ILP-style regularizers are kept disabled (weight
zero) to keep the loss exactly the four terms displayed in
\hyperref[alg:unified]{algorithm}~\ref{alg:unified}; adding these auxiliary
terms with non-zero weights was found to destabilize scenario-2 training on the
non-\(1\)-unambiguous operator \(\sigma = \times\).

\paragraph*{Metrics}
Let \(\mathcal{B}\) be the validation set of \(N_{\text{val}}\) bags, each of size \(M\).
For bag \(b\), write \(s_b\) for the observed bag label,
\(\hat{s}_b = \arg\max_t \hat{p}_t\) for the predicted bag label (the argmax of
the combined bag distribution \(\hat{p}\) of
\hyperref[alg:unified]{algorithm}~\ref{alg:unified}), \(y_{b,i}\) for the (held-out) gold instance label at
position \(i\) and \(\hat{y}_{b,i} = \arg\max_y p_\theta(y \mid x_{b,i})\)
(the per-instance CP probability of \hyperref[alg:unified]{algorithm}~\ref{alg:unified})
for the predicted instance label. The reported metrics are:
\begin{enumerate}
\item[---] \textbf{BagAcc} \(= \tfrac{1}{N_{\text{val}}} \sum_b \mathbf{1}\{\hat{s}_b = s_b\}\):
 correctness at the weak-supervision level.
\item [---]\textbf{InstAcc} \(= \tfrac{1}{N_{\text{val}}M} \sum_b \sum_{i=1}^{M} \mathbf{1}\{\hat{y}_{b,i} = y_{b,i}\}\):
 alignment with latent instance-level labels.
\item [---]\textbf{Constraint} \(= \tfrac{1}{N_{\text{val}}} \sum_b \mathbf{1}\{\sigma(\hat{y}_{b,1}, \ldots, \hat{y}_{b,M}) = s_b\}\):
 fraction of bags whose hard predicted instance labels satisfy the induced
 TP rule.
\item [---]\textbf{Bag--Constraint} \(= \mathrm{BagAcc}-\mathrm{Constraint}\): a positive
 gap signals bag-level success that is \emph{not} structurally supported by
 the induced TP rule (e.g.\ an OP\(\leftrightarrow\)TP shortcut).
\item [---]\textbf{Bag--Inst} \(= \mathrm{BagAcc}-\mathrm{InstAcc}\): a positive
 gap signals misalignment between bag predictions and instance semantics
 (e.g.\ correct sums from wrong digits).
\end{enumerate}

\paragraph*{Evaluation pathways} BagAcc and Constraint are computed on
different pipelines. BagAcc evaluates the combined distribution
\(\hat{p}\) of \hyperref[alg:unified]{algorithm}~\ref{alg:unified}, which
mixes the TP-marginalized CP prediction with the OP candidate
distribution at mixture weight \(w_{\text{OP}}\) (initialized at
\(0.3\) and learned jointly in the scenario-2 runs; fixed at \(0.3\) in
the scenario-1 drivers; \hyperref[alg:unified]{algorithm}~\ref{alg:unified}
states the general three-weight combination, of which the evaluated
configuration is the two-component form with \(w_{\text{TP}} = 0\)); since the candidate set contains the true
label, the OP component places probability \(1/3\) on \(s_b\) and
assists \(\hat{s}_b\). Constraint applies \(\sigma\) to the hard
per-instance CP predictions with no OP contribution. A positive
Bag--Constraint gap can therefore reflect either an
OP\(\leftrightarrow\)TP shortcut or the OP-mixture assistance in
\(\hat{s}_b\); reading it together with InstAcc separates the two. In
the `no TP' and `OP only' ablations the bag prediction is produced by a
direct bag classifier over the CP outputs, without access to the
candidate distribution, so those rows are not assisted by OP at
evaluation.

For TP inference (experiment~1), the following per-operator
quantities are further defined. Given the candidate operator set \(\Sigma\) and predicted instance
labels \(\hat{y}_{b,i}\), let
\(E^+(\sigma) = \{b \in \mathcal{B} : \sigma(\hat{y}_{b,1}, \ldots, \hat{y}_{b,M}) = s_b\}\)
and \(E^-(\sigma) = \mathcal{B} \setminus E^+(\sigma)\) for \(\sigma \in \Sigma\).
The TP-discovery phase trains a soft weight vector \(w \in \Delta(\Sigma)\)
over the candidate operators on the training split
(the softmax-normalized learnable operator scores \(g_\phi\) of
\hyperref[alg:unified]{algorithm}~\ref{alg:unified}); the selected operator
is \(\sigma^\star = \arg\max_{\sigma \in \Sigma} w_\sigma\). In the
strong-CP regime this coincides with maximizing the discrete gap
\(|E^+(\sigma)| - |E^-(\sigma)|\); under random CP the two argmaxes can
differ, and \(\sigma^\star\) as reported is always the weight argmax.
\begin{enumerate}
\item[---] \textbf{OpSelAcc} \(= \mathbf{1}\{\sigma^\star = \sigma_{\text{true}}\}\),
 averaged over seeds.
\item [---]\textbf{ConsScore}\((\sigma) = |E^+(\sigma)| - |E^-(\sigma)| \in [-N_{\text{val}}, N_{\text{val}}]\),
 the discrete ILP objective at \(\sigma\), evaluated on \(\mathcal{B}\).
 ConsScore\((\sigma_{\text{true}})\) is reported.
\item [---]\textbf{ConsRatio}\((\sigma) = |E^+(\sigma)| / N_{\text{val}} \in [0, 1]\), reported at
 \(\sigma_{\text{true}}\).
\item [---]\textbf{RatioMargin} \(= \mathrm{ConsRatio}(\sigma_{\text{true}}) - \max_{\sigma \in \Sigma \setminus \{\sigma_{\text{true}}\}} \mathrm{ConsRatio}(\sigma)\):
 separation between the true operator's constraint-satisfaction rate and
 the best competing operator's. The margin is negative when a competing
 operator explains more validation bags than the true one, as happens
 under random CP and under extreme OP corruption
 (\hyperref[tab:op_noise]{table}~\ref{tab:op_noise}).
\end{enumerate}

For CP inference (experiment~2), \textbf{F1m} is additionally reported, the
macro-averaged F1 score across the ten digit classes, i.e.\
\(\mathrm{F1m} = \tfrac{1}{10} \sum_{j=0}^{9} F_{1,j}\) with
\(F_{1,j} = 2\,\mathrm{Prec}_j\,\mathrm{Rec}_j / (\mathrm{Prec}_j + \mathrm{Rec}_j)\),
where \(\mathrm{Prec}_j\) and \(\mathrm{Rec}_j\) are the per-class
precision and recall of the hard predictions \(\hat{y}_{b,i}\) against the
gold instance labels \(y_{b,i}\), and the convention \(F_{1,j} = 0\)
whenever class \(j\) has no true positives (in particular
\(\mathrm{Prec}_j := 0\) when no instance is predicted as \(j\)).

\subsection{Full model and ablations}
The full model is first evaluated alongside three ablations that
selectively remove semantic components: `no TP' (the TP is removed from the loss), `no CP' (the CP is removed from the loss) and `OP only' (the three
semantic terms \(\mathcal{L}_{\text{con}}, \mathcal{L}_{\text{cp}}\text{ and }
\mathcal{L}_{\text{ctp}}\) are set to zero, leaving only the bag-level
cross-entropy \(\mathcal{L}_{\text{bag}}\) on the OP supervision). All
models are trained using identical OP supervision.

\paragraph*{Scope of comparisons}
The goal is not to provide an exhaustive MI-PLL benchmark, since many existing
methods assume a fixed aggregation model or different supervision regimes.
Instead, controlled variants are compared that isolate the semantic effect of
explicit CP/TP/OP structure: OP-only learning (which can fit bag labels
spuriously) and ablations that remove CP or TP. This design tests, within
this controlled setting, whether explicit semantic coupling is needed for
reliable recovery under this weak-supervision construction.

\hyperref[tab:unified_ablation]{Table} {\ref{tab:unified_ablation}} reports results across aggregation operators
over 20 random seeds. The full model achieves bag accuracy and constraint
satisfaction of approximately 0.96 on every operator with near-zero
Bag--Constraint gap, consistent with bag-level success being structurally
supported by the induced TP rule rather than by spurious correlations
(the two evaluation pathways of \S4a agree to within \(0.008\)). The
instance-level picture diverges between operators: addition and
multiplication both reach instance accuracy above 0.97, while XOR shows a
markedly bimodal instance accuracy across seeds, a per-seed
collapse pattern that is invisible to bag accuracy and constraint
satisfaction. This bimodality is an empirical analogue of the failure mode captured by
Wang \textit{et~al.}'s lemma~1 \cite{wang2024learning} for non-\(M\)-unambiguous operators
(\hyperref[app:wang:implications]{appendix C.4}).

The ablations expose where each semantic component contributes.
Removing CP creates an OP$\leftrightarrow$TP shortcut: bag accuracy
survives partially (\(0.48\) on \(+\) and \(0.53\) on \(\times\)) while
instance accuracy drops to the digit-level chance baseline (\(0.10\), the
\(1/10\) marginal across ten digit classes) and the Bag--Constraint gap
grows to 0.34--0.47, indicating that the model satisfies bag labels
through the TP rule alone without recovering instance semantics.
Removing TP eliminates this shortcut, but it also removes the mechanism
through which OP supervision can structure CP, so bag accuracy drops to
the bag-level marginal baseline across operators (near-uniform chance for
\(+\) and \(\oplus\); the modal-class rate for \(\times\), whose
zero-absorber makes \(s = 0\) appear in \(19/100\) ordered pairs). The
OP-only baseline, trained with the three semantic terms (constraint,
CP-confidence and CP--TP consistency) zeroed and only the bag-level
cross-entropy active, also collapses to the same baseline, confirming that weak
supervision alone, without the CP--TP coupling, provides too little
gradient signal to recover useful predictions. The pattern is consistent
across operators: every per-operator failure mode is diagnosable through
the combination of bag accuracy and the two gap metrics rather than
through bag accuracy alone.

\begin{table}
\tabcolsep8.8pt
\tbl{Full model ablations across aggregation operators (mean $\pm$ s.d., 20 seeds).}{%
\begin{tabular}{@{\Tabind}lllllll@{\Tabind}}
\multicolumn{1}{l@{}}{\cellcolor{rstaclr}\tabhead{\hspace*{-3pt}op\hspace{3pt}}} &
\multicolumn{1}{l@{}}{\cellcolor{rstaclr}\tabhead{Ablation}} &
\multicolumn{1}{l@{}}{\cellcolor{rstaclr}\tabhead{BagAcc}} &
\multicolumn{1}{l@{}}{\cellcolor{rstaclr}\tabhead{InstAcc}} &
\multicolumn{1}{l@{}}{\cellcolor{rstaclr}\tabhead{Constraint}} &
\multicolumn{1}{l@{}}{\cellcolor{rstaclr}\tabhead{Bag--Constraint}} &
\multicolumn{1}{l@{}}{\cellcolor{rstaclr}\tabhead{Bag--Inst}} \\
\colrule
$+$ &full & 0.961 $\pm $0.010 & 0.978 $\pm $0.005 & 0.956 $\pm $0.010 & 0.005$ \pm $0.004 & --0.017 $\pm $0.006 \\
\midrule
$+$ &no CP & 0.481 $\pm $0.018 & 0.100$ \pm $0.009 & 0.009$ \pm $0.005 & 0.471$ \pm $0.017 &\hspace*{3pt} 0.381$ \pm$ 0.022\hspace*{-3pt} \\
\midrule
$+$ &no TP & 0.096$ \pm $0.017 & 0.102 $\pm $0.011 & 0.041 $\pm $0.022 & 0.056$ \pm $0.027 & --0.006 $\pm $0.020 \\
\midrule
$+$ &OP only & 0.096 $\pm $0.017 & 0.098$ \pm $0.014 & 0.057$ \pm $0.030 & 0.039$ \pm$ 0.034 & --0.002$ \pm $0.019 \\
\midrule
$\times$ & full & 0.957 $\pm $0.013 & 0.975 $\pm $0.006 & 0.957$ \pm$ 0.012 & 0.000 $\pm$ 0.003 & --0.018$ \pm$ 0.007 \\
\midrule
$\times$ & no CP & 0.528$ \pm $0.023 & 0.100$ \pm $0.009 & 0.190 $\pm $0.017 & 0.338$ \pm $0.022 &\hspace*{3pt} 0.428 $\pm $0.021\hspace*{-3pt} \\
\midrule
$\times$ & no TP & 0.190 $\pm $0.017 & 0.102$ \pm$ 0.010 & 0.042$ \pm $0.054 & 0.148$ \pm$ 0.053 &\hspace*{3pt} 0.088$ \pm $0.021\hspace*{-3pt} \\
\midrule
$\times$ & OP only & 0.190$ \pm $0.017 & 0.097 $\pm $0.013 & 0.060 $\pm $0.069 & 0.130 $\pm $0.069 &\hspace*{3pt} 0.093 $\pm $0.019\hspace*{-3pt} \\
\midrule
$\oplus$ & full & 0.962 $\pm $0.008 & 0.439$ \pm $0.497 & 0.955$ \pm $0.010 & 0.007$ \pm $0.005 &\hspace*{3pt} 0.523$ \pm $0.500\hspace*{-3pt} \\
\midrule
$\oplus$ & no CP & 0.441$ \pm $0.016 & 0.100$ \pm$ 0.009 & 0.104$ \pm $0.014 & 0.337$ \pm$ 0.016 &\hspace*{3pt} 0.341 $\pm $0.018\hspace*{-3pt} \\
\midrule
$\oplus$ & no TP & 0.100$ \pm$ 0.014 & 0.104 $\pm $0.014 & 0.104 $\pm $0.014 & --0.004 $\pm $0.016 & --0.004$ \pm $0.020 \\
\midrule
$\oplus$ & OP only  & 0.115 $\pm $0.037 & 0.098$ \pm $0.022 & 0.105$ \pm $0.016 & 0.009$ \pm $0.037 &\hspace*{3pt} 0.017 $\pm $0.051\hspace*{-3pt} \\
\botrule
\end{tabular}}\label{tab:unified_ablation}
\end{table}

\paragraph*{Operationalization of the semantics}
\hyperref[alg:unified]{Algorithm}~\ref{alg:unified} summarizes how the OP--CP--TP semantics are
operationalized in the $\alpha$-ILP-style differentiable pipeline. The algorithm is provided
for clarity rather than as a novel training procedure. The purpose is to make
explicit how neural perception (CP), symbolic hypothesis evaluation (TP) and
weak supervision (OP) interact within a single learning loop under logical
constraints.

The specific loss terms and optimization details are not the focus of this
work; these terms serve only to enforce the semantic constraints defined by OP, CP
and TP. The claims here concern the necessity of these semantic dependencies rather
than the particular training dynamics of the pipeline.

\begin{figure}
\centering
\includegraphics[width=0.95\linewidth]{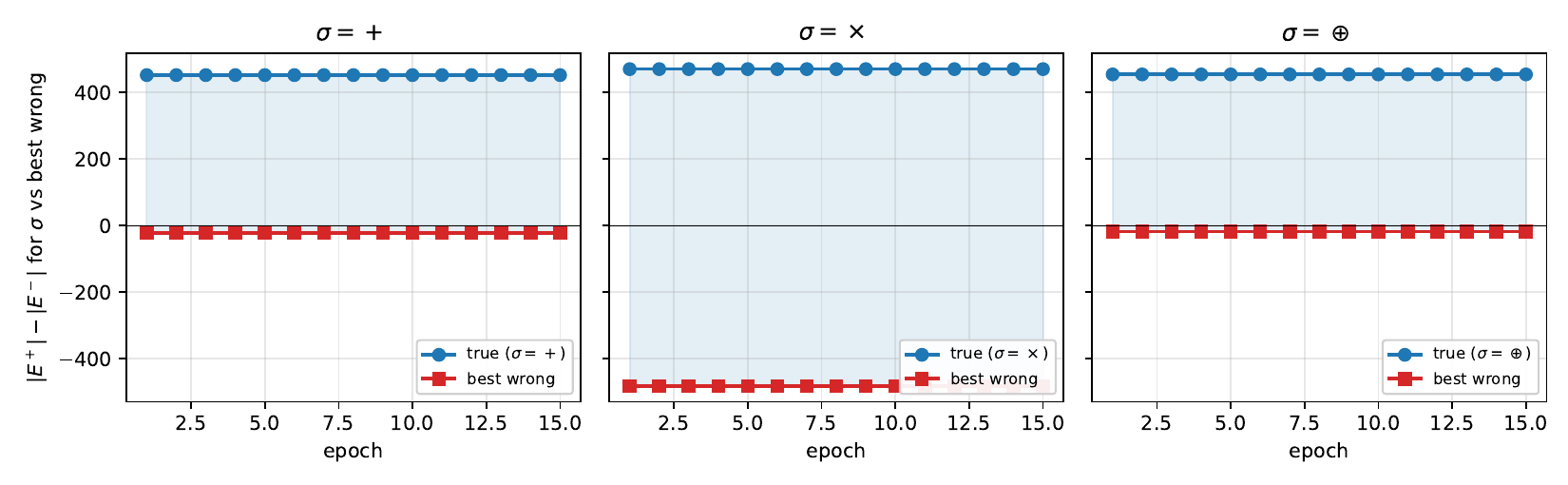}
\caption{Per-epoch \(|E^+|-|E^-|\) separation between the true operator \(\sigma\) and the best competing operator, for one representative seed per operator. The true-operator scores are positive and stable from epoch~1 for all three operators; the best competing operator's scores are negative throughout (substantially so for \(\sigma = \times\)). The shaded region marks the separation \(|E^+_{\text{true}}|-|E^-_{\text{true}}| - (|E^+_{\text{best wrong}}|-|E^-_{\text{best wrong}}|)\), the margin by which the true operator maximizes the discrete ILP objective.}
\label{fig:eplus_eminus}
\end{figure}

\paragraph*{Surrogate relationship to the \(|E^+|-|E^-|\) objective} The
bag-level loss \(\mathcal{L}_{\text{bag}} = \sum_b -\log \hat{p}_{s_b}\)
is a smooth surrogate for the discrete count of bag-level
mispredictions. Writing \(E^\pm\) here for the partition induced by the
hard bag prediction \(\hat{s}_b = \arg\max_t \hat{p}_t\) of the same
distribution \(\hat{p}\) that appears in the loss
(\hyperref[app:proofs]{appendix~B}, whose per-bag bound holds for any
distribution in \(\Delta(S)\), in particular the mixed \(\hat{p}\)),
\(|E^-| \leq \mathcal{L}_{\text{bag}} / \log 2\) as derived below. For a
deterministic classifier this partition coincides with the
\(\sigma(\mathbf{f}(\mathbf{x})) = s\) partition of
\hyperref[sec:learning-scenarios]{\S3}; for a
probabilistic classifier the two can differ, and the Constraint metric
of \S4a tracks the latter.
\hyperref[prop:surrogate]{Proposition}~\ref{prop:surrogate} establishes that whenever the
chosen hard-argmax prediction \(\hat{s}_b\) (under a fixed tie-breaking
convention) is incorrect on bag \(b\),
\(\hat{p}_{s_b} \leq \tfrac{1}{2}\) and hence
\(-\log \hat{p}_{s_b} \geq \log 2\). Summing over bags yields the
one-sided inequality
\(\mathcal{L}_{\text{bag}} \geq (\log 2)\,|E^-|\), equivalently
\(|E^-| \leq \mathcal{L}_{\text{bag}} / \log 2\). Cross-entropy is not
monotone in \(|E^-|\) in the other direction: a lower
\(\mathcal{L}_{\text{bag}}\) does not imply a lower \(|E^-|\) in
general, but minimizing \(\mathcal{L}_{\text{bag}}\) minimizes an
upper bound on \(|E^-|\) and, equivalently, maximizes the induced lower
bound \(N - 2\mathcal{L}_{\text{bag}}/\log 2\) on
\(|E^+| - |E^-| = N - 2|E^-|\). The
remaining terms \(\mathcal{L}_{\text{con}}\), \(\mathcal{L}_{\text{cp}}\) and
\(\mathcal{L}_{\text{ctp}}\) add structural regularization that encourages the
CP-derived and TP-derived bag predictions to agree, without altering this
surrogate relationship. The discrete \(|E^+| - |E^-|\) gap is also tractable
to track explicitly across training (counting, per candidate operator,
the validation bags on which \(\sigma\) applied to the hard per-instance
predictions matches \(s_b\)); doing so produces the trajectories of
\hyperref[fig:eplus_eminus]{figure} {\ref{fig:eplus_eminus}}, in which the true \(\sigma\) is separated
from the best competing operator by a large positive gap from epoch~1
onwards (the figure shows all three operators; for
\(\sigma = \oplus\) the bag-level separation holds regardless of
instance-level collapse, since XOR-equivariant relabellings preserve bag
consistency).

\begin{algorithm}
\caption{Operationalization of OP--CP--TP semantics in the $\alpha$-ILP-style differentiable pipeline.}
\label{alg:unified}
\begin{algorithmic}[1]
\Require Bags $\mathbf{x}=(x_1,\ldots,x_M)$, weak labels $s$, candidate operators $\Sigma$
\Require Neural CP model $f_\theta$, optional learnable TP $g_\phi$, OP candidate sets
\Require Loss weights $\lambda_{\text{bag}}, \lambda_{\text{con}}, \lambda_{\text{cp}}, \lambda_{\text{ctp}}$
\For{each minibatch of bags}
 \State \textbf{CP prediction:} compute per-instance probabilities $p_\theta(y_i \mid x_i)$
 \State \textbf{OP distribution:} build $p_{\text{OP}} \in \Delta(S)$, uniform over each bag's candidate label set
 \State \textbf{TP/Operator scores (soft):}
 \State \hspace{1em} compute bag-level scores for each $\sigma\in\Sigma$ and/or $g_\phi$
 \State \textbf{CP-derived bag distribution:} $\hat{p}_{\text{CP}} \in \Delta(S)$ from $p_\theta(y_i\mid x_i)$ marginalized through the TP operator
 \State \textbf{TP-derived bag distribution:} $\hat{p}_{\text{TP}} \in \Delta(S)$ from operator scores
 \State \textbf{Combined bag distribution (soft combine,} with mixture weights $w_{\text{OP}}, w_{\text{CP}}, w_{\text{TP}} \geq 0$ summing to one\textbf{):}
 $\hat{p} = w_{\text{OP}}\,p_{\text{OP}} + w_{\text{CP}}\,\hat{p}_{\text{CP}} + w_{\text{TP}}\,\hat{p}_{\text{TP}} \;\in\; \Delta(S)$
 \State \textbf{Losses} (with $\hat{p}_s$, $\hat{p}_{\text{TP},s}$ the $s$-th components of $\hat{p}$ and $\hat{p}_{\text{TP}}$; $p_\theta(y\mid x_i)$ the per-instance CP probability):
 $\mathcal{L}_{\text{bag}} = -\log \hat{p}_s$ \\
 $\mathcal{L}_{\text{con}} = -\log \hat{p}_{\text{TP},s}$ \\
 $\mathcal{L}_{\text{cp}} = -\frac{1}{M}\sum_{i=1}^{M} \max_{y \in Y} p_\theta(y \mid x_i)$ \\
 $\mathcal{L}_{\text{ctp}} = \|\hat{p}_{\text{TP}} - \hat{p}_{\text{CP}}\|_2^2$ \\
 \State \textbf{Total Loss:}
 $\mathcal{L} = \lambda_{\text{bag}}\mathcal{L}_{\text{bag}} +
 \lambda_{\text{con}}\mathcal{L}_{\text{con}} +
 \lambda_{\text{cp}}\mathcal{L}_{\text{cp}} +
 \lambda_{\text{ctp}}\mathcal{L}_{\text{ctp}}$
 \State Update $\theta$ (and $\phi$ if TP is learnable) via backpropagation
\EndFor
\end{algorithmic}
\end{algorithm}

\subsection{Experiment 1: transition predicate inference from observed predicate and classifier predicate}
This experiment corresponds to scenario~1 in the semantic formulation: inferring
the TP given OP and CP. A soft weight vector over the enumerated candidate
operators is trained by the four-term loss, and the operator with the
largest learned weight is selected (\hyperref[sec:experiments-and-evaluation]{\S4a}).

To test sensitivity to perceptual reliability, \textbf{strong CP}
(a classifier pre-trained in the scenario-2 configuration, i.e.\ under
the true operator, before the operator weights are trained) is compared
against \textbf{random CP} (untrained). With strong CP, scenario~1
therefore measures operator \emph{identification} given a classifier
consistent with the true operator; the random-CP condition probes
discovery without that assistance.
Results are shown in \hyperref[tab:exp1_tp]{table}~\ref{tab:exp1_tp}. With strong CP, TP inference
succeeds for all three operators, achieving 100\% operator selection
accuracy on every seed and large positive consistency scores. The
\(\times\) row reaches reliable operator selection (OpSelAcc \(1.000\)) only after the scenario-1
training configuration is matched to the four-term loss of
\hyperref[alg:unified]{algorithm}~\ref{alg:unified} (the four-term weights
$(\lambda_{\text{bag}}, \lambda_{\text{con}}, \lambda_{\text{cp}},
\lambda_{\text{ctp}}) = (1.0, 1.0, 0.05, 0.5)$, squared
$\ell_2$ distance for $\mathcal{L}_{\text{ctp}}$, and joint
CP--TP optimization rather than the alternating-phase variant); an
alternative configuration with non-zero auxiliary entropy and ILP weights
left the TP-inference driver in the non-\(1\)-unambiguous
multiplicative zero-attractor (\hyperref[app:wang:implications]{appendix C.4})
and collapsed operator selection to chance. Under random CP, the consistency signal collapses across all three operators (ConsRatio drops to chance, and ConsScore becomes large and negative). Operator selection behaves differently by operator: for \(+\) and \(\oplus\) the argmax still recovers the true \(\sigma\) (OpSelAcc \(\geq 0.95\)), since the relative ranking of operators survives even when none of the operators fits well, whereas for \(\times\) the multiplicative zero-attractor disrupts even this ranking and OpSelAcc drops to \(0.200 \pm 0.410\). These results
show that, in this setup, absolute consistency collapses under random
CP, and for \(\times\) even the operator ranking fails; the training
configuration also needs to be matched to the operator's unambiguity
profile.

\begin{table}
\tabcolsep13.9pt
\tbl{Experiment 1: TP inference under strong versus random CP (mean $\pm$ s.d., 20 seeds).}{%
\begin{tabular}{@{\Tabind}llllll@{\Tabind}}
\multicolumn{1}{l@{}}{\cellcolor{rstaclr}\tabhead{\hspace*{-7pt}op\hspace*{7pt}}} &
\multicolumn{1}{l@{}}{\cellcolor{rstaclr}\tabhead{CP setting}} &
\multicolumn{1}{l@{}}{\cellcolor{rstaclr}\tabhead{OpSelAcc}} &
\multicolumn{1}{l@{}}{\cellcolor{rstaclr}\tabhead{ConsRatio}} &
\multicolumn{1}{l@{}}{\cellcolor{rstaclr}\tabhead{ConsScore}} &
\multicolumn{1}{l@{}}{\cellcolor{rstaclr}\tabhead{RatioMargin}} \\
\colrule
$+$ &strong & 1.000 $\pm $0.000 & 0.949 $\pm $0.010 & 448.90 $\pm $10.31 & 0.456$ \pm$ 0.020 \\
\midrule
$+$ &random & 1.000$ \pm $0.000 & 0.056$ \pm $0.022 & --443.80$ \pm $22.50 & 0.013$ \pm$ 0.020 \\
\midrule
$\times$ & strong & 1.000$ \pm$ 0.000 & 0.953$ \pm$ 0.015 & 452.90$ \pm$ 15.10 & 0.930$ \pm $0.016 \\
\midrule
$\times$ & random & 0.200 $\pm $0.410 & 0.045 $\pm $0.053 & --454.60$ \pm $53.21 & --0.062$ \pm$ 0.048 \\
\midrule
$\oplus$ & strong & 1.000$ \pm $0.000 & 0.949$ \pm $0.010 & 448.70$ \pm $10.33 & 0.448$ \pm$ 0.017 \\
\midrule
$\oplus$ & random & 0.950$ \pm$ 0.224 & 0.089$ \pm$ 0.015 & --410.80$ \pm $14.93 & 0.036 $\pm $0.022 \\
\botrule
\end{tabular}}\label{tab:exp1_tp}
\end{table}

\paragraph*{Background-knowledge size} Wang \textit{et~al.}'s theorem~5 \cite{wang2024learning} is a
\emph{sample-complexity} statement: a larger candidate-operator
class \(\mathcal{G}\) increases the number of partially-labelled
samples needed to reach a given classification risk, with the
dependence captured by the term \(d_{\mathcal{G}}\). Optimization
epochs are not the same quantity as sample complexity, so theorem~5 cannot
be tested directly with the training-loop measurements. A
related empirical trend is reported instead: scenario~1 was run with
\(|\Sigma|\) varied from \(1\) (oracle) to \(5\) (true operator plus
four distractors, including \(\max\) and \(\min\)). TP identification
succeeds with OpSelAcc \(= 1.00\) at every \(|\Sigma|\), with the
convergence epoch growing modestly from \(1\) to \(2\) as more
candidates are added. This trend is consistent with the qualitative
direction of Wang \textit{et~al.}'s theorem~5 \cite{wang2024learning} (more candidates require more
data/effort), without being a formal test of the bound. Full numbers
are reported in \hyperref[app:bk_sweep]{appendix A.2}.

\subsection{Experiment 2: classifier predicate inference from observed predicate and fixed transition predicate}
This experiment corresponds to scenario~2: inferring instance-level classifier
assignments given OP and a fixed TP. \textbf{Strong
CP} (trained under OP constraints) is compared against \textbf{weak CP} (minimally
trained: two epochs on 200 bags).

\hyperref[tab:exp2_cp]{Table} \ref{tab:exp2_cp} shows that weak CP fails to recover semantic structure
on every operator: bag accuracy stalls around \(0.45\)--\(0.50\), a level attributable
largely to the OP-candidate component of the evaluation mixture (which
places probability \(1/3\) on the true label; \S4a) rather than to
recovered structure, while constraint satisfaction stays below \(0.22\)
and instance accuracy stays below \(0.34\). OP
supervision alone is insufficient to recover meaningful instance-level
predictions, regardless of the operator. Strong CP, in contrast, achieves
bag accuracy and constraint satisfaction of approximately 0.96 on all
three operators. Instance accuracy splits along the Wang \textit{et~al.}\
\cite{wang2024learning} unambiguity profile (\hyperref[app:wang:unambig]{appendix C.2}): for the
\(M\)-unambiguous operators \(\sigma = +\) and \(\sigma = \times\),
instance accuracy reaches 0.97--0.98 in lockstep with bag accuracy. For
\(\sigma = \oplus\), which is not \(M\)-unambiguous, instance accuracy
is sharply bimodal across the 20 seeds: approximately half of the seeds
recover digits with instance accuracy above 0.9 and the other half show
instance accuracy near 0 (aggregate \(0.439 \pm 0.497\)) despite preserved
bag and constraint accuracy on every seed. This pattern is consistent
with collapse to an XOR-equivariant relabelling (mechanism detailed in
\hyperref[app:wang:implications]{appendix C.4}); the per-seed mapping from confusion matrices is not separately
verified, so the verified claim is that
the observation is explainable by such a relabelling rather than that
each collapsed seed is one. The bimodality is also consistent with the
failure mode that Wang \textit{et~al.}'s lemma~1 \cite{wang2024learning} establishes existentially for
non-\(M\)-unambiguous operators (\hyperref[app:wang:implications]{appendix C.4}):
lemma~1 proves the existence of a data distribution \(\mathcal{D}\) and
a classifier with partial risk zero and classification risk one; an
XOR-equivariant relabelling would realize this configuration on the
actual MNIST distribution rather than on Wang \textit{et~al.}'s \cite{wang2024learning} adversarial
construction.

\begin{table}
\tabcolsep6.4pt
\tbl{Experiment 2: CP quality affects bag/constraint performance (mean $\pm$ s.d., 20 seeds). XOR instance-accuracy and F1 are bimodal across seeds; Bag--Constraint and Bag--Inst are derived from the BagAcc, Constraint and InstAcc columns.}{%
\begin{tabular}{@{\Tabind}llllllll@{\Tabind}}
\multicolumn{1}{l@{}}{\cellcolor{rstaclr}\tabhead{op}} &
\multicolumn{1}{l@{}}{\cellcolor{rstaclr}\tabhead{CP setting}} &
\multicolumn{1}{l@{}}{\cellcolor{rstaclr}\tabhead{BagAcc}} &
\multicolumn{1}{l@{}}{\cellcolor{rstaclr}\tabhead{Constraint}} &
\multicolumn{1}{l@{}}{\cellcolor{rstaclr}\tabhead{InstAcc}} &
\multicolumn{1}{l@{}}{\cellcolor{rstaclr}\tabhead{F1m}} &
\multicolumn{1}{l@{}}{\cellcolor{rstaclr}\tabhead{Bag--Constraint}} &
\multicolumn{1}{l@{}}{\cellcolor{rstaclr}\tabhead{Bag--Inst}} \\
\colrule
$+$ &strong & 0.961 $\pm $0.010 & 0.956 $\pm$ 0.010 & 0.978 $\pm$ 0.005 & 0.977 $\pm$ 0.005 & 0.005 & --0.017 \\
\midrule
$+$ &weak & 0.500$ \pm $0.025 & 0.079 $\pm$ 0.031 & 0.178 $\pm$ 0.061 & 0.090 $\pm$ 0.049 & 0.421 & 0.322 \\
\midrule
$\times$ & strong & 0.957$ \pm $0.013 & 0.957 $\pm$ 0.012 & 0.975 $\pm$ 0.006 & 0.974 $\pm$ 0.006 & 0.000 & --0.018 \\
\midrule
$\times$ & weak & 0.491 $\pm$ 0.110 & 0.218 $\pm$ 0.069 & 0.335 $\pm$ 0.097 & 0.263 $\pm$ 0.097 & 0.272 & 0.155 \\
\midrule
$\oplus$ & strong & 0.962$ \pm $0.008 & 0.955 $\pm$ 0.010 & 0.439 $\pm$ 0.497 & 0.439 $\pm$ 0.497 & 0.007 & 0.523 \\
\midrule
$\oplus$ & weak & 0.447 $\pm$ 0.019 & 0.117 $\pm$ 0.019 & 0.140 $\pm$ 0.044 & 0.057 $\pm$ 0.041 & 0.331 & 0.307 \\
\botrule
\end{tabular}}\label{tab:exp2_cp}
\end{table}

\subsection{Observed predicate quality sensitivity (noise study)}
Robustness to weak supervision quality is analysed by progressively corrupting
OP bag labels with increasing levels of noise and evaluating TP inference. In
this setting, noise is injected directly into the training OPs
(validation labels remain clean), while
the underlying instance labels and aggregation rules remain unchanged. TP inference is
evaluated using \textbf{ConsRatio} and \textbf{RatioMargin}. ConsRatio measures the fraction of validation bags whose hard CP predictions are consistent with the true operator (\S4a); RatioMargin measures the gap to the best competing operator.

\begin{figure}
\centering
\includegraphics[width=0.75\linewidth]{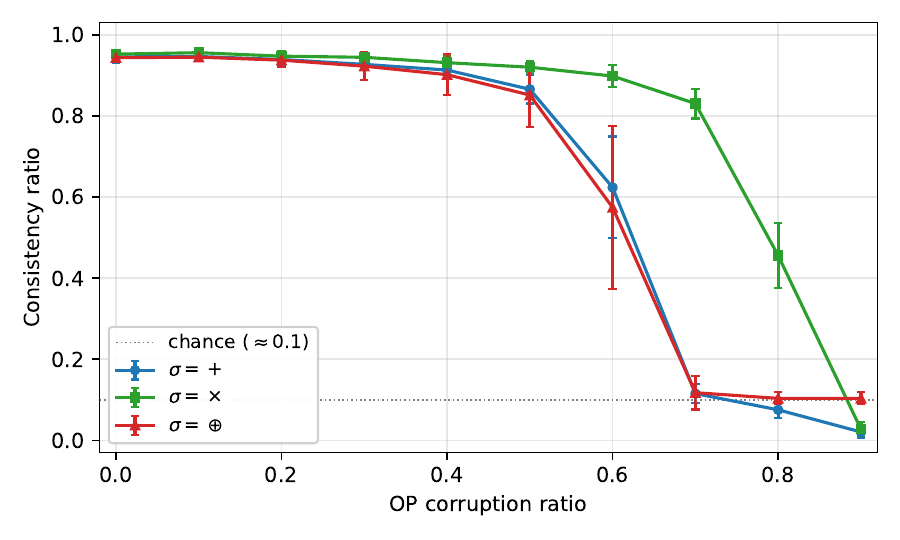}
\caption{OP-noise sensitivity for TP inference. Consistency ratio as a function of the OP-corruption ratio, mean $\pm$ s.d. over 10 seeds. Addition and XOR plateau above \(0.85\) up to noise \(0.5\) and transition sharply to chance between \(0.5\) and \(0.7\). Multiplication is empirically more robust: it stays at or above \(0.83\) up to noise \(0.7\) and drops sharply between \(0.7\) and \(0.8\). This operator-by-operator asymmetry is an empirical observation, not a prediction of the clean-label learnability theory of Wang \textit{et~al.}~\cite{wang2024learning}; a plausible structural explanation is the asymmetric pre-image distribution of \(\times\) on the digit range.}
\label{fig:op_noise}
\end{figure}

All three operators exhibit a plateau in the consistency ratio
followed by a sharp transition, but the transition location differs
across operators. Addition and XOR maintain ConsRatio above \(0.85\)
up to OP-corruption ratio \(0.5\), then drop sharply between \(0.5\)
and \(0.7\) before falling to or below chance. By contrast, multiplication
stays at or above \(0.83\) up to noise \(0.7\) and drops sharply between
\(0.7\) and \(0.8\) (\hyperref[fig:op_noise]{figure}~\ref{fig:op_noise}). The transition in
either regime is abrupt rather than gradual, consistent with the
bag-level evidence flipping from informative to
anti-informative.
This behaviour is consistent
with the view of weak supervision as a structured logical constraint
that the symbolic layer can denoise within the operative
regime. The operator-by-operator difference in noise tolerance is
consistent with the recent theoretical analysis of Tsamoura
\textit{et~al.}~\cite{tsamoura2025imbalances}, which establishes class-specific risk
bounds in MI-PLL (their proposition~3.1) that show the symbolic
component \(\sigma\) can induce significant per-class learning
imbalances even when the hidden-label marginal is uniform. The
mechanism Tsamoura \textit{et~al.}\ identify, namely an asymmetric pre-image distribution
under \(\sigma\), is also present for multiplication on the digit
range: the bag label \(0\) has an unusually large pre-image because any
pair containing a zero maps to \(0\). This makes corrupted labels of
that form especially easy for the multiplicative operator to
rationalize and provides a plausible structural explanation for the
observed noise plateau. Their illustrative \(\sigma = \mathrm{max}\)
example exhibits the same general kind of pre-image asymmetry, though in
the opposite direction. The multiplication noise
plateau is therefore interpreted as empirically related to the pre-image-asymmetry mechanism
Tsamoura \textit{et~al.}\ \cite{tsamoura2025imbalances} identify under clean weak-label structure, rather than
as a contingent empirical artefact. Tsamoura \textit{et~al.}'s \cite{tsamoura2025imbalances} theory does not
formally cover OP-corruption regimes, so this is a structural motivation
rather than a formal corollary of those bounds. The numerical results for this
subsection are reported in \hyperref[app:op_noise]{appendix A.1}, with a
detailed mapping to their framework in \hyperref[app:tsamoura]{appendix D}.

\paragraph*{Scalability and computational cost}
The hypothesis space is intentionally restricted to a small operator set in
order to isolate semantic dependencies. TP selection requires evaluating
candidate hypotheses for consistency, and this introduces overhead compared to
purely neural baselines. Runtime scales with the number of bags, the number of
candidate operators and the number of constraint instances. This is regarded as
the cost of enforcing explicit semantic consistency under weak supervision, and
more expressive operator libraries and more search-efficient ILP
strategies are left to future work.

\paragraph*{Why not an oracle}
Unlike an oracle program that hard-codes the true operator, the present setting must
infer TP from weak supervision and noisy neural CP outputs; residual errors are
therefore expected owing to ambiguity in OP labels and local optima in joint
neuro-symbolic optimization.

\section{Related work}
The closest theoretical companions to this work are Wang \textit{et~al.}~\cite{wang2024learning} and the recent work of Tsamoura \textit{et~al.}~\cite{tsamoura2025imbalances}. Wang \textit{et~al.}\ \cite{wang2024learning} supply sample-complexity bounds for MI-PLL under unambiguity conditions on the aggregation operator $\sigma$. Tsamoura \textit{et~al.}\ \cite{tsamoura2025imbalances} extend this analysis to class-specific risk bounds, show that $\sigma$ can cause per-class learning imbalances even under uniform marginals, and propose linear-programming (LP)-based training-time mitigation together with a robust optimal-transport test-time mitigation algorithm. Their contribution is statistical and algorithmic; \hyperref[app:tsamoura]{appendix D} discusses how those imbalance-mitigation techniques compose with the present framework.

Several neuro-symbolic frameworks combine logic with neural learning to handle structured supervision: Scallop~\cite{scallop}, DeepProbLog~\cite{DBLP:journals/corr/abs-1907-08194, deepproblog-approx} (built on Sato's distribution semantics~\cite{DBLP:conf/iclp/Sato95}), DeepStochLog~\cite{DeepStochlog22}, NeurASP~\cite{neurasp}, NeuroLog~\cite{neurolog}, SATNet~\cite{wang2019satNet}, \MakeLowercase{Logic Tensor Networks}~\cite{DBLP:journals/corr/SerafiniG16}, \MakeLowercase{Logical Neural Networks}~\cite{sen22}, \MakeLowercase{Neural Theorem Provers}~\cite{rocktaschel2017ntp}, $\partial$Forth~\cite{pmlr-v70-bosnjak17a}, \MakeLowercase{Semantic Loss}~\cite{pmlr-v80-xu18h}, DL2~\cite{dl2}, abductive learning~\cite{ABL}, neural-guided program synthesis~\cite{kalyan2018neuralguided} and related variants~\cite{iclr2023, feldstein2023principled}. These systems integrate symbolic constraints through different mechanisms (probabilistic logic programs, soft satisfaction and abduction). The experimental pipeline here follows the differentiable-ILP design of $\alpha$-ILP~\cite{shindo2023alpha}, representing neural predicates and scoring symbolic hypotheses by differentiable consistency.

Single-instance partial label learning has been studied extensively~\cite{EM-PLL, cour2011, structured-prediction-pll, progressive-pll, deep-naive-pll, weighthed-loss-pll, instance-dependent-pll, noisy-pll, learnability-pll}, including transition-matrix-based analyses for noisy supervision~\cite{noisy-multiclass-learning, proper-losses-pll, JMLR:corrupted-learning, proper-losses-pll-pkdd}; this work targets the multi-instance setting~\cite{demipl2023, zhang2024mipl, zhang2025fastmipl}, where bag-level supervision aggregates latent instance labels; multi-instance learning is reviewed by~Foulds \&\ Frank \cite{mil-review1}, with theoretical learnability bounds for bag-level supervision developed by~Sabato \textit{et~al.}\ \cite{sabato-mil-bags}. Learning from label proportions~\cite{scott2020llp, zhang2022llp} is a related bag-level weak-supervision paradigm in which bags carry class-proportion annotations rather than operator-aggregated labels. Other approaches to weak supervision include relaxed supervision~\cite{relaxed-supervision}, structured prediction with paired examples~\cite{spigot1, spigot2, word-problem1, word-problem2, paired-examples} and combinatorial solvers integrated into neural architectures~\cite{mipaal, comboptnet}. Weak supervision in natural language processing~\cite{pmlr-v48-raghunathan16, hu-etal-2016-harnessing} is a separate major line focused on text-domain structure.

\section{Discussion}
The CP/TP/OP-coupled model consistently outperforms the no-CP, no-TP and OP-only ablations across operators (\hyperref[tab:unified_ablation]{table}~\ref{tab:unified_ablation}), indicating that structured symbolic guidance both improves bag-level prediction and reveals semantic failure modes that bag accuracy alone would hide.

At the theoretical level, this work builds on Wang \textit{et~al.}~\cite{wang2024learning}, whose
theorem~5 establishes empirical-risk-minimization learnability under an \emph{unknown} transition
class \(\mathcal{G}\), conditional on the unambiguity of \(\mathcal{G}\)
(definition~5 there) and the \emph{\(r\)-bounded} assumption:
every classifier \(f\) in the search space \(\mathcal{F}\) satisfies the
class-conditional accuracy lower bound \(\mathbb{P}([f](X) = y \mid Y = y) \geq r\)
for some \(r > 0\) and every label \(y \in Y\). The empirical
observation that TP discovery (scenario~1) requires a competent CP,
documented in the strong-CP versus random-CP comparison of
experiment~1 (\hyperref[tab:exp1_tp]{table}~\ref{tab:exp1_tp}), is consistent with this
condition: a cold-start CP has near-random hard predictions \([f](x)\),
so the class-conditional hard accuracy
\(\mathbb{P}([f](X) = y \mid Y = y)\) can be near chance (in
expectation \(\approx 1/c\), where \(c = 10\) is the number of digit
classes, under symmetric random initialization) or
even zero on some classes that the cold-start argmax happens to never
output; the zero case falls below every positive \(r\), and the
near-chance case below any \(r\) exceeding \(1/c\), so the
\(r\)-bounded premise fails at the strengths the
learnability bounds need. Scenario-2 pre-training lifts the
pre-trained CP's class-conditional hard accuracy above \(1/c\) on
non-collapsed seeds, giving an empirical analogue of Wang \textit{et~al.}'s
\cite{wang2024learning} \(r\)-bounded condition for the specific CP that is then used. The
\(r\)-bounded assumption is a property of the entire search class
\(\mathcal{F}\), not of an individual classifier, so this correspondence
is operational rather than a literal class-level verification. The same observation has a sharper reading
under the class-specific risk framework of Tsamoura
\textit{et~al.}~\cite{tsamoura2025imbalances}: a cold-start CP also has near-trivial
partial risk \(\mathcal{R}^{01}_{\textsf{P}}(f; \sigma) \approx 1\), at
which Tsamoura \textit{et~al.}'s \cite{tsamoura2025imbalances} class-specific bound \(\Phi_{\sigma, j}\)
(Tsamoura \textit{et~al.}'s proposition~3.1; growth curves shown in Tsamoura \textit{et~al.}'s figure~2) is
uninformative: by construction \(\Phi_{\sigma, j}(R_{\textsf{P}}) \leq 1\),
matching the trivial bound \(R_j(f) \leq 1\) on every class. Scenario-2
pre-training lowers the partial risk into the regime where the bound
becomes non-trivial and recovers a per-class profile that TP discovery
can use to distinguish candidate operators. The formal mappings
are detailed in \hyperref[app:wang]{appendices C} and~\hyperref[app:tsamoura]{D}.

The present work builds on recent advances in differentiable ILP, including the foundational $\partial$ILP system of Evans \&\ Grefenstette~\cite{Evans18} and the $\alpha$-ILP system~\cite{shindo2023alpha}, which integrates logical clause learning with neural perception predicates. Extensions such as Shindo \textit{et~al.}~\cite{shindo21}, Sen \textit{et~al.}~\cite{sen22}, Krishnan \textit{et~al.}~\cite{krishnan21} and Gao \textit{et~al.}~\cite{dforl2024} have expanded this line of research by enabling clause-level control, supporting first-order logic, and introducing predicate invention and recursive rule learning. However, a formal semantics for MI-PLL within such systems remains underdeveloped. The semantics formalization here provides one step towards filling that gap by making explicit the structural assumptions that ILP systems can exploit in MI-PLL settings.

Three challenges remain for future work. Cross-domain robustness requires the framework's behaviour to remain consistent across variations in image representations, which can vary significantly across datasets. The expressivity of the transition logic is constrained: both scenarios assume a finite candidate set of symbolic aggregation operators, and the dependence on hand-engineered background knowledge limits adaptability; extending ILP to support flexible rule discovery and learned background knowledge would broaden the framework beyond the current candidate operator set. Delayed supervision is another challenge: the current formulation assumes that OP labels are immediately available for every bag, whereas many practical settings provide feedback that is delayed or only conditionally available, requiring extension to time-indexed or deferred logical feedback in OP.

The approach does not modify the classifier's architecture; the symbolic constraints are introduced through the loss function during training. The framework also enables diagnostic comparisons between the classifier's outputs and the ILP-derived consistent predictions, which can expose semantic failure modes such as reasoning shortcuts~\cite{marconato23, bears2024}. The gap metrics reported in \hyperref[sec:experiments-and-evaluation]{\S}\ref{sec:experiments-and-evaluation} (Bag--Constraint and Bag--Inst) operationalize this kind of diagnostic check.

Although this paper addresses only two specific scenarios, TP inference given an available CP and CP inference given a fixed TP, these provide a starting point for a broader ILP-based treatment of MI-PLL. Future work will generalize this to cases with latent or partially known predicates, explore more complex supervision structures and expand empirical testing across diverse datasets. A direct extension is to compose the CP/TP/OP semantics with the class-specific imbalance-mitigation algorithms of Tsamoura \textit{et~al.}~\cite{tsamoura2025imbalances}: the marginal-estimation procedure, LP pseudolabel assignment and the CAROT test-time score adjustment (confidence adjustment via robust semi-constrained optimal transport) plug into the CP layer and could be used to address the XOR bimodal seed-collapse, and the post-plateau collapse in multiplication under OP noise. The mapping is detailed in \hyperref[app:tsamoura]{appendix D}.

\section{Conclusion}
This paper introduced a semantic framework for MI-PLL\ based on three auxiliary predicates: the CP, the TP and the OP. Two inductive tasks were formalized within this framework, inferring TP given an available CP and inferring CP given a fixed TP, and the framework was evaluated using an $\alpha$-ILP-style differentiable pipeline as the backend on MI-PLL tasks constructed from MNIST. Across the operators $+$, $\times$ and $\oplus$, the experiments expose operator-dependent failure modes that bag accuracy alone would hide, in line with the unambiguity conditions of Wang \textit{et~al.}\ \cite{wang2024learning} and the class-specific risk framework of Tsamoura \textit{et~al.} \cite{tsamoura2025imbalances}. Extending the framework to larger operator libraries, delayed supervision and direct integration with class-specific mitigation algorithms remains future work.

\bigskip

\paragraph*{Data accessibility} \mbox{}\\ This article has no additional data. The code that supports the findings is available at \url{https://github.com/nuuoe/ILP_MI-PLL}.

\appendix
\renewcommand{\thesection}{A}
\setcounter{subsection}{0}
\section*{Appendix A. Additional experimental results}
\renewcommand{\thesubsection}{\thesection.\arabic{subsection}.}
\renewcommand{\thesubsubsection}{\thesubsection\arabic{subsection}.}

\subsection{Observed predicate noise sensitivity}
\label{app:op_noise}

\hyperref[tab:op_noise]{Table} \ref{tab:op_noise} reports the full per-noise-level TP inference results that underlie the OP-noise discussion in \hyperref[sec:experiments-and-evaluation]{\S}\ref{sec:experiments-and-evaluation}. Results are shown as mean $\pm$ standard deviation over 10 seeds, with the scenario-1 training configuration matched to the \hyperref[alg:unified]{algorithm}~\ref{alg:unified} four-term loss. All three operators exhibit a plateau-then-cliff pattern: addition and XOR transition between noise \(0.5\) and \(0.7\), while multiplication transitions between \(0.7\) and \(0.8\). The clean-label sample-complexity bounds of~Wang \textit{et~al.}\ \cite{wang2024learning} do not directly model OP corruption and therefore do not predict this asymmetry in the transition location; a structural interpretation in terms of Tsamoura \textit{et~al.}'s \cite{tsamoura2025imbalances} class-specific framework (which is a clean-weak-label theory under known \(\sigma\), not an OP-noise theory) is given in \hyperref[app:tsamoura]{appendix D}.

\begin{table}
\tabcolsep37.4pt
\tbl{OP-noise sensitivity: ConsRatio and RatioMargin under OP corruption (mean $\pm$ s.d., 10 seeds).}{%
\begin{tabular}{@{\Tabind}llll@{\Tabind}}
\multicolumn{1}{l@{}}{\cellcolor{rstaclr}\tabhead{\hspace*{-29pt}op\hspace{29pt}}} &
\multicolumn{1}{l@{}}{\cellcolor{rstaclr}\tabhead{Noise}} &
\multicolumn{1}{l@{}}{\cellcolor{rstaclr}\tabhead{ConsRatio }}&
\multicolumn{1}{l@{}}{\cellcolor{rstaclr}\tabhead{RatioMargin}} \\
\colrule
$+$ & 0.0 & 0.945 $\pm$ 0.015 & 0.456 $\pm$ 0.012 \\
\midrule
$+$ & 0.1 & 0.946 $\pm$ 0.012 & 0.457 $\pm$ 0.014 \\
\midrule
$+$ & 0.2 & 0.939 $\pm$ 0.013 & 0.452 $\pm$ 0.012 \\
\midrule
$+$ & 0.3 & 0.927 $\pm$ 0.015 & 0.446 $\pm$ 0.015 \\
\midrule
$+$ & 0.4 & 0.913 $\pm$ 0.020 & 0.438 $\pm$ 0.013 \\
\midrule
$+$ & 0.5 & 0.865 $\pm$ 0.035 & 0.406 $\pm$ 0.021 \\
\midrule
$+$ & 0.6 & 0.623 $\pm$ 0.126 & 0.276 $\pm$ 0.070 \\
\midrule
$+$ & 0.7 & 0.115 $\pm$ 0.023 & 0.016 $\pm$ 0.006 \\
\midrule
$+$ & 0.8 & 0.075 $\pm$ 0.020 & 0.010 $\pm$ 0.005 \\
\midrule
$+$ & 0.9 & 0.021 $\pm$ 0.016 & 0.001 $\pm$ 0.004 \\
\midrule
$\times$ & 0.0 & 0.952 $\pm$ 0.011 & 0.930 $\pm$ 0.011 \\
\midrule
$\times$ & 0.1 & 0.956 $\pm$ 0.008 & 0.933 $\pm$ 0.011 \\
\midrule
$\times$ & 0.2 & 0.947 $\pm$ 0.012 & 0.924 $\pm$ 0.014 \\
\midrule
$\times$ & 0.3 & 0.944 $\pm$ 0.013 & 0.921 $\pm$ 0.015 \\
\midrule
$\times$ & 0.4 & 0.931 $\pm$ 0.016 & 0.907 $\pm$ 0.020 \\
\midrule
$\times$ & 0.5 & 0.920 $\pm$ 0.014 & 0.896 $\pm$ 0.014 \\
\midrule
$\times$ & 0.6 & 0.898 $\pm$ 0.028 & 0.875 $\pm$ 0.028 \\
\midrule
$\times$ & 0.7 & 0.830 $\pm$ 0.037 & 0.806 $\pm$ 0.038 \\
\midrule
$\times$ & 0.8 & 0.456 $\pm$ 0.079 & 0.416 $\pm$ 0.084 \\
\midrule
$\times$ & 0.9 & 0.028 $\pm$ 0.017 & --0.088 $\pm$ 0.056 \\
\midrule
$\oplus$ & 0.0 & 0.943 $\pm$ 0.010 & 0.448 $\pm$ 0.014 \\
\midrule
$\oplus$ & 0.1 & 0.945 $\pm$ 0.007 & 0.449 $\pm$ 0.014 \\
\midrule
$\oplus$ & 0.2 & 0.937 $\pm$ 0.017 & 0.444 $\pm$ 0.022 \\
\midrule
$\oplus$ & 0.3 & 0.922 $\pm$ 0.034 & 0.434 $\pm$ 0.023 \\
\midrule
$\oplus$ & 0.4 & 0.901 $\pm$ 0.051 & 0.427 $\pm$ 0.028 \\
\midrule
$\oplus$ & 0.5 & 0.851 $\pm$ 0.079 & 0.395 $\pm$ 0.034 \\
\midrule
$\oplus$ & 0.6 & 0.574 $\pm$ 0.201 & 0.278 $\pm$ 0.091 \\
\midrule
$\oplus$ & 0.7 & 0.118 $\pm$ 0.041 & 0.093 $\pm$ 0.012 \\
\midrule
$\oplus$ & 0.8 & 0.104 $\pm$ 0.015 & 0.104 $\pm$ 0.015 \\
\midrule
$\oplus$ & 0.9 & 0.104 $\pm$ 0.015 & 0.104 $\pm$ 0.015 \\
\botrule
\end{tabular}}\label{tab:op_noise}
\end{table}

\subsection{Background-knowledge size}
\label{app:bk_sweep}

To probe the effect of background-knowledge size $|B|$ on TP discovery
empirically, scenario~1 was run with the candidate-operator set
$\Sigma$ varied from $|\Sigma| = 1$ (only the true operator) to
$|\Sigma| = 5$ (the set $\{+, \times, \oplus, \max, \min\}$, which
contains the true operator); \(|B|\) is measured here by \(|\Sigma|\),
the number of candidate operators the background theory encodes.
This corresponds operationally to varying $d_{\mathcal{G}}$ in
theorem~5 of Wang \textit{et~al.}\ \cite{wang2024learning}
(\hyperref[app:wang:bk]{appendix C.6}). Results, averaged over 10 seeds, are
reported in \hyperref[tab:bk_sweep]{table}~\ref{tab:bk_sweep}. OpSelAcc
is as in \S4a; ConvRate is the fraction of seeds whose true-operator
weight exceeds \(0.9\) during training; ConvEp is the first epoch at
which this occurs; margin denotes the
operator-weight margin
\(w_{\sigma_{\text{true}}} - \max_{\sigma \neq \sigma_{\text{true}}} w_\sigma\)
of the learned operator weights at the final epoch (at \(|B| = 1\) the
competitor set is empty and the margin is defined as
\(w_{\sigma_{\text{true}}}\)).

Operator-selection accuracy holds at 1.00 across $|B| \in \{1, \ldots, 5\}$
for all three operators, confirming that TP identification succeeds under
strong CP for every candidate-set size in this range. The convergence
epoch grows from 1.0 at $|B| = 1$ to approximately 2.0 at $|B| \geq 3$ for $+$ and
$\oplus$, and similarly for $\times$ (with a slightly slower
transition at $|B| = 3$, mean $1.7 \pm 0.5$, indicating that some
seeds still converge in one epoch at \(|B| = 3\)). Enlarging the candidate set does not change the final
operator selection in any of the runs. This is read as the empirical
\emph{analogue} of the qualitative direction of Wang \textit{et~al.}'s
theorem~5: smaller $\mathcal{G}$ yields smaller sample complexity in their setting, and smaller optimization-epoch counts are observed here.
The two quantities (theorem~5's \(m_{\textsf{P}}\) and the epoch
count) are not the same; the convergence-epoch trend is a related
empirical signal, not a formal test of the bound.

\begin{table}
\tabcolsep23.5pt
\tbl{TP-discovery convergence as a function of background-knowledge size $|B|$ (mean $\pm$ s.d., 10 seeds).}{%
\begin{tabular}{@{\Tabind}llllll@{\Tabind}}
\multicolumn{1}{l@{}}{\cellcolor{rstaclr}\tabhead{\hspace*{-15.5pt}op\hspace*{15.5pt}}} &
\multicolumn{1}{l@{}}{\cellcolor{rstaclr}\tabhead{$|B|$}} &
\multicolumn{1}{l@{}}{\cellcolor{rstaclr}\tabhead{OpSelAcc}} &
\multicolumn{1}{l@{}}{\cellcolor{rstaclr}\tabhead{ConvRate }}&
\multicolumn{1}{l@{}}{\cellcolor{rstaclr}\tabhead{ConvEp}} &
\multicolumn{1}{l@{}}{\cellcolor{rstaclr}\tabhead{Margin}} \\
\colrule
$+$ &1 & 1.00 & 1.00 & 1.0 $\pm$ 0.0 & 1.000 $\pm$ 0.000 \\
\midrule
$+$ &2 & 1.00 & 1.00 & 1.0 $\pm$ 0.0 & 0.999 $\pm$ 0.000 \\
\midrule
$+$ &3 & 1.00 & 1.00 & 1.9 $\pm$ 0.3 & 0.999 $\pm$ 0.000 \\
\midrule
$+$ &4 & 1.00 & 1.00 & 2.0 $\pm$ 0.0 & 0.999 $\pm$ 0.000 \\
\midrule
$+$ &5 & 1.00 & 1.00 & 2.0 $\pm$ 0.0 & 0.998 $\pm$ 0.001 \\
\midrule
$\oplus$ & 1 & 1.00 & 1.00 & 1.0 $\pm$ 0.0 & 1.000 $\pm$ 0.000 \\
\midrule
$\oplus$ & 2 & 1.00 & 1.00 & 1.0 $\pm$ 0.0 & 0.999 $\pm$ 0.000 \\
\midrule
$\oplus$ & 3 & 1.00 & 1.00 & 2.0 $\pm$ 0.0 & 0.999 $\pm$ 0.001 \\
\midrule
$\oplus$ & 4 & 1.00 & 1.00 & 2.0 $\pm$ 0.0 & 0.999 $\pm$ 0.000 \\
\midrule
$\oplus$ & 5 & 1.00 & 1.00 & 2.0 $\pm$ 0.0 & 0.999 $\pm$ 0.000 \\
\midrule
$\times$ & 1 & 1.00 & 1.00 & 1.0 $\pm$ 0.0 & 1.000 $\pm$ 0.000 \\
\midrule
$\times$ & 2 & 1.00 & 1.00 & 1.0 $\pm$ 0.0 & 0.999 $\pm$ 0.000 \\
\midrule
$\times$ & 3 & 1.00 & 1.00 & 1.7 $\pm$ 0.5 & 0.999 $\pm$ 0.000 \\
\midrule
$\times$ & 4 & 1.00 & 1.00 & 2.0 $\pm$ 0.0 & 0.999 $\pm$ 0.000 \\
\midrule
$\times$ & 5 & 1.00 & 1.00 & 2.0 $\pm$ 0.0 & 0.999 $\pm$ 0.000 \\
\botrule
\end{tabular}}\label{tab:bk_sweep}
\end{table}

\renewcommand{\thesection}{B}
\setcounter{section}{0}
\section*{Appendix B. Formal statements and proofs}
\label{app:proofs}

This appendix restates, with explicit assumptions, the load-bearing arithmetic and
combinatorial claims made throughout the main text and provides concise
proofs.

\paragraph*{Setting and notation} Fix a finite label set \(Y\), an instance
space \(X\), a bag size \(M \geq 1\) and a finite bag-label space \(S\).
The true aggregation rule \(\sigma \colon Y^M \to S\) and the latent
data-generating process \(D \colon X^M \to S\) are total functions related
by \(D(\mathbf{x}) = \sigma(g(x_1), \ldots, g(x_M))\), where
\(g \colon X \to Y\) is the (latent) per-instance gold-label map;
throughout the appendix, \(s^{(b)} = D(\mathbf{x}^{(b)})\) denotes
the observed bag label without reference to the latent factorization. A
weakly-labelled \emph{dataset} is a finite list
\(\mathcal{D} = ((\mathbf{x}^{(b)}, s^{(b)}))_{b=1}^{N}\) with
\(\mathbf{x}^{(b)} \in X^M\) and \(s^{(b)} = D(\mathbf{x}^{(b)})\). A
\emph{hypothesis} is a pair \(H = (f, \sigma)\), where \(\sigma \colon Y^M \to S\)
is drawn from a finite candidate set \(\Sigma\), and \(f\) is either a
deterministic classifier \(X \to Y\) or, more generally, a probabilistic
classifier \(\tilde{f} \colon X \to \Delta(Y)\), where \(\Delta(Y)\) is the
probability simplex over \(Y\).

Every hypothesis induces a \emph{soft bag distribution}
\(\hat{p}(H, \mathbf{x}) \in \Delta(S)\) by marginalizing the operator
\(\sigma\) over the per-instance distributions:
\[
\hat{p}(H, \mathbf{x})_t
\;=\; \sum_{\mathbf{y} \in Y^M,\; \sigma(\mathbf{y}) = t}
 \prod_{i=1}^{M} \tilde{f}(x_i)_{y_i},
\qquad t \in S.
\]
The associated \emph{hard prediction} is
\(\hat{s}(H, \mathbf{x}) := \arg\max_{t \in S} \hat{p}(H, \mathbf{x})_t\),
with ties broken by any fixed rule. For a deterministic classifier,
\(\tilde{f}\) is a Dirac mass and \(\hat{p}(H, \mathbf{x})\) reduces to a
Dirac at \(\sigma(f(x_1), \ldots, f(x_M))\). For a probabilistic
classifier, the argmax-of-marginal prediction used here need not
coincide with \(\sigma\) applied to the per-instance argmaxes (the
Constraint metric of \S4a); \hyperref[prop:surrogate]{proposition}~\ref{prop:surrogate}
concerns the former, i.e.\ the bag-level prediction scored by
\(\mathcal{L}_{\mathrm{bag}}\). Define the hypothesis-dependent
partition
\[
E^+(H) \;=\; \{(\mathbf{x}, s) \in \mathcal{D} \,:\, \hat{s}(H, \mathbf{x}) = s\}\text{ and}
\qquad
E^-(H) \;=\; \{(\mathbf{x}, s) \in \mathcal{D} \,:\, \hat{s}(H, \mathbf{x}) \neq s\}.
\]
Bags are assumed pairwise distinct, so the two sets partition
\(\mathcal{D}\); equivalently, \(E^\pm(H)\) may be read as index sets
over \(\{1, \ldots, N\}\). The standard information-theoretic conventions \(0 \log 0 = 0\)
and \(-\log 0 = +\infty\) are adopted.

\paragraph*{Notation conventions used throughout the main text} The worked examples and formal statements use the following two conventions throughout:
\begin{enumerate}
\item[---] \emph{Sample form for training data.} A training example with bag \((x_1, \ldots, x_M)\) and observed weak label \(s\) is written compactly as the tuple \((x_1, \ldots, x_M, s)\); this is the same datum that the atom \(\text{OP}((x_1, \ldots, x_M), s)\) records. Sample form is used when enumerating training data; the formal predicate form is used inside constraint structures, theorem statements and proofs.
\item [---]\emph{Operator subscripts.} The \(\text{TP}\), the intermediate-interpretation structure \(\mathbb{I}\) and the hypothesis \(\mathbb{H}\) all carry a subscript indicating the aggregation operator. When the operator is \emph{fixed and given as part of the problem setup} (as in scenario~2 of \hyperref[sec:learning-scenarios]{\S}\ref{sec:learning-scenarios}, where \(\sigma\) is provided), the symbol is written directly: \(\text{TP}_+, \text{TP}_\times, \text{TP}_\oplus, \text{TP}_C\), and likewise \(\mathbb{I}_+, \mathbb{I}_\times, \mathbb{I}_\oplus, \mathbb{I}_C\) and \(\mathbb{H}_+, \mathbb{H}_\times, \mathbb{H}_\oplus, \mathbb{H}_C\). When the operator is \emph{hypothesized, generic, unknown or being searched over} (as in scenario~1, where \(\sigma\) is inferred), \(\text{TP}_\sigma\), \(\mathbb{I}_\sigma\) and \(\mathbb{H}_\sigma\) are written.
\end{enumerate}

\begin{lemma}[partition exhaustiveness]
\label{lem:partition}
For every hypothesis \(H\) and dataset \(\mathcal{D}\) of size \(N\),
\[
|E^+(H)| + |E^-(H)| \;=\; N
\quad\text{and}\quad
E^+(H) \cap E^-(H) \;=\; \emptyset.
\]
\end{lemma}

\begin{proof}
For each \((\mathbf{x}, s) \in \mathcal{D}\), the predicate
\(\hat{s}(H, \mathbf{x}) = s\) is decidable and has a unique truth value;
the two cases partition \(\mathcal{D}\) into disjoint subsets \(E^+(H)\)
and \(E^-(H)\) whose union is \(\mathcal{D}\). The cardinality equality
follows by additivity, and disjointness is immediate from the definition.
\end{proof}

\begin{proposition}[objective reduction]
\label{thm:reduction}
Let the ILP objective be \(\max_H (|E^+(H)| - |E^-(H)|)\), where
\(|E^+(H)|\text{ and }|E^-(H)| \in \mathbb{Z}_{\geq 0}\) are the cardinalities of
the partition induced by \(H\) on a dataset of \(N\) bags. Then for every
hypothesis \(H\),
\[
|E^+(H)| - |E^-(H)| \;=\; 2\,|E^+(H)| - N.
\]
Consequently the map \(H \mapsto |E^+(H)| - |E^-(H)|\) is a strictly
increasing affine function of \(|E^+(H)|\), so
\[
\arg\max_{H} \bigl(|E^+(H)| - |E^-(H)|\bigr)
\;=\;\arg\max_{H} |E^+(H)|,
\]
that is, the ILP objective is equivalent to maximizing bag-level
classification accuracy under the decomposition \(H = (f, \sigma)\).
\end{proposition}
\begin{proof}
By \hyperref[lem:partition]{lemma}~\ref{lem:partition}, \(|E^-(H)| = N - |E^+(H)|\). Substituting
yields \(|E^+(H)| - |E^-(H)| = |E^+(H)| - (N - |E^+(H)|) = 2|E^+(H)| - N\).
The right-hand side has slope \(+2\) in \(|E^+(H)|\) and constant offset
\(-N\), so the argmax over \(H\) of the gap coincides with the argmax of
\(|E^+(H)|\). Bag-level classification accuracy on \(\mathcal{D}\) is
defined as \(|E^+(H)|/N\); for fixed \(N\), maximizing \(|E^+(H)|\) is the
same as maximizing accuracy.
\end{proof}

\begin{proposition}[cross-entropy surrogate bound]
\label{prop:surrogate}
Let \(p \in \Delta(S)\), \(s \in S\), and let \(\hat{s}\) be any chosen
hard prediction satisfying \(p_{\hat{s}} \geq p_t\) for all \(t \in S\).
If \(\hat{s} \neq s\), then \(p_s \leq \tfrac{1}{2}\).
Consequently, the per-bag cross-entropy loss
\(\ell(p, s) := -\log p_s\) satisfies
\[
\hat{s} \neq s
\;\Longrightarrow\;
\ell(p, s) \geq \log 2,
\]
under the convention \(-\log 0 = +\infty\). Summing over the dataset, the
bag-level loss
\(\mathcal{L}_{\mathrm{bag}}(H) := \sum_{(\mathbf{x}, s) \in \mathcal{D}}
\ell(\hat{p}(H, \mathbf{x}), s)\) satisfies
\[
\mathcal{L}_{\mathrm{bag}}(H) \;\geq\; (\log 2) \cdot |E^-(H)|.
\]
\end{proposition}
\begin{proof}
Suppose the chosen hard prediction is \(j \neq s\); then by definition,
\(p_j \geq p_s\). Since \(p\) is a probability distribution, all entries
are non-negative and \(\sum_{t \in S} p_t = 1\). Because \(j \neq s\),
the terms \(p_j\) and \(p_s\) are two distinct entries of this sum, so
\(p_j + p_s \leq \sum_{t \in S} p_t = 1\), which gives
\(p_j \leq 1 - p_s\). Combining,
\(p_s \leq p_j \leq 1 - p_s\), hence \(2 p_s \leq 1\) and
\(p_s \leq \tfrac{1}{2}\).

For the loss bound, \(-\log\) is strictly decreasing on \((0, 1]\), and
the convention \(-\log 0 = +\infty\) extends the inequality to the
boundary: for any \(p_s, q \in [0, 1]\) with \(p_s \leq q\) and \(q > 0\),
\(-\log p_s \geq -\log q\). Applied to \(p_s \leq \tfrac{1}{2}\)
this yields \(-\log p_s \geq -\log \tfrac{1}{2} = \log 2\) (with
\(+\infty \geq \log 2\) when \(p_s = 0\)).

For the dataset bound, observe that \(\ell(p, s) = -\log p_s \geq 0\) for
every \(p \in \Delta(S)\) and every \(s\) (because \(p_s \leq 1\)), so the
contributions from \(E^+(H)\) are non-negative. By the previous step, each
contribution from \(E^-(H)\) is at least \(\log 2\). Therefore
\[
\mathcal{L}_{\mathrm{bag}}(H) \;=\;
\sum_{b \in E^+(H)} \ell(\hat{p}(H, \mathbf{x}^{(b)}), s^{(b)})
\;+\;
\sum_{b \in E^-(H)} \ell(\hat{p}(H, \mathbf{x}^{(b)}), s^{(b)})
\;\geq\; 0 + (\log 2) \cdot |E^-(H)|.
\]
\end{proof}

\paragraph*{Remark (mixture instantiation)} The per-bag inequality of
\hyperref[prop:surrogate]{proposition}~\ref{prop:surrogate} holds for any
\(p \in \Delta(S)\), so the dataset-level bound applies verbatim when each
bag's distribution is the mixed \(\hat{p}\) of
\hyperref[alg:unified]{algorithm}~\ref{alg:unified} rather than the pure
CP-marginal \(\hat{p}(H, \mathbf{x})\) defined above, with \(E^\pm(H)\)
read as the partition induced by the argmax of that same mixed
distribution.

\begin{lemma}[Boolean C asymmetry]
\label{lem:booleanC}
Define \(\sigma_C(y_1, y_2, y_3) = (y_1 \wedge y_2) \vee (y_1 \wedge y_3)\)
for \(y_1, y_2, y_3 \in \{0, 1\}\) (the Boolean function underlying the
predicate \(\mathrm{TP}_C\) of \hyperref[sec:scenario2]{\S3b}). Then \(\sigma_C\) is invariant
under the transposition \(y_2 \leftrightarrow y_3\) but \emph{not} invariant
under the transposition \(y_1 \leftrightarrow y_2\).
\end{lemma}
\begin{proof}
The first claim is the chain
\(\sigma_C(y_1, y_3, y_2) = (y_1 \wedge y_3) \vee (y_1 \wedge y_2)
= (y_1 \wedge y_2) \vee (y_1 \wedge y_3) = \sigma_C(y_1, y_2, y_3)\),
using commutativity of \(\vee\). For the second, take
\((y_1, y_2, y_3) = (0, 1, 1)\). Direct evaluation gives
\(\sigma_C(0, 1, 1) = (0 \wedge 1) \vee (0 \wedge 1) = 0\) and
\(\sigma_C(1, 0, 1) = (1 \wedge 0) \vee (1 \wedge 1) = 1\), so the two
inputs related by the \(y_1 \leftrightarrow y_2\) transposition produce
different outputs.
\end{proof}

\begin{lemma}[XOR pre-image cardinality]
\label{lem:xorpreimage}
Let \(Y = \{0, 1, \ldots, 9\}\) and let \(\oplus\) denote bitwise XOR.
For every \(s \in \{0, 1, \ldots, 15\}\),
\[
|\{(a, b) \in Y \times Y \,:\, a \oplus b = s\}| \;\geq\; 2.
\]
\end{lemma}
\begin{proof}
If \(s = 0\), the pre-image contains every diagonal pair \((a, a)\) for
\(a \in Y\) (since \(a \oplus a = 0\)), giving cardinality \(10 \geq 2\).
Suppose \(s \neq 0\). The map \(a \mapsto a \oplus s\) is an involution of
the non-negative integers (since XOR is self-inverse) and has no fixed
points on \(s \neq 0\) (because \(a \oplus s = a\) implies \(s = 0\)).
Consequently, whenever \((a, b) \in Y \times Y\) is a pre-image (i.e.\
\(a \oplus b = s\)), the swapped pair \((b, a)\) is also in \(Y \times Y\)
and is a pre-image; it is distinct from \((a, b)\) because \(a \neq b\)
(as \(a = b\) would force \(s = a \oplus a = 0\)).

It therefore suffices to exhibit a single pre-image pair for each
\(s \in \{1, \ldots, 15\}\). For \(s \in \{1, \ldots, 9\}\), the pair
\((0, s) \in Y \times Y\) satisfies \(0 \oplus s = s\). For
\(s \in \{10, 11, 12, 13, 14, 15\}\), the pair \((s \oplus 9, 9)\) lies in
\(Y \times Y\) because
\(10 \oplus 9 = 3,\ 11 \oplus 9 = 2,\ 12 \oplus 9 = 5,\
13 \oplus 9 = 4,\ 14 \oplus 9 = 7,\ 15 \oplus 9 = 6\)
are all in \(\{0, \ldots, 9\}\), and by construction
\((s \oplus 9) \oplus 9 = s\). Combining with the involution argument
gives at least two distinct pre-images for each such \(s\), completing the
proof.
\end{proof}

\paragraph*{Remarks on usage in the main text}
\hyperref[lem:partition]{Lemma}~\ref{lem:partition} and \hyperref[thm:reduction]{proposition}~\ref{thm:reduction} jointly formalize
the abductive-partition remark of
\hyperref[sec:learning-scenarios]{\S3} and provide
the precise sense in which the stated ILP objective reduces to bag-level
classification accuracy under the \((f, \sigma)\) decomposition.
\hyperref[prop:surrogate]{Proposition}~\ref{prop:surrogate} justifies the surrogate-relationship
paragraph in \hyperref[sec:experiments-and-evaluation]{\S}\ref{sec:experiments-and-evaluation}: minimizing the
soft bag-level cross-entropy minimizes an upper bound on the discrete
misclassification count \(|E^-|\); explicitly,
\(|E^-| \leq \mathcal{L}_{\text{bag}} / \log 2\). \hyperref[lem:booleanC]{Lemma}~\ref{lem:booleanC} confirms the partial
permutation-invariance caveat in~\hyperref[sec:scenario2]{\S3b}. \hyperref[lem:xorpreimage]{Lemma}~\ref{lem:xorpreimage}
makes precise the structural ambiguity of \(\sigma = \oplus\) on the digit
range: every feasible XOR bag label has at least two distinct
instance-pair pre-images, in line with the empirical observation that no
bag label has a unique pre-image on \(Y\).

\renewcommand{\thesection}{C}
\setcounter{theorem}{0}
\section*{Appendix C. Mapping to learnability theory}
\label{app:wang}

The present paper builds on the multi-instance partial label learnability
framework of Wang \textit{et~al.}\ \cite{wang2024learning}. This appendix supplies
the precise mapping between Wang \textit{et~al.}'s \cite{wang2024learning} formal apparatus and the
CP/TP/OP semantics developed in the main text. The relevant theorems are restated,
the unambiguity conditions are verified for each operator used
in the experiments, the empirical interpretations supported are spelled out, and
what the ILP-based extension adds on top of these results is made explicit.

\setcounter{subsection}{0}
\subsection{Notation}
\label{app:wang:notation}

Wang \textit{et~al.}\ \cite{wang2024learning} study \emph{MI-PLL}: pairs \((\mathbf{x}, s)\) with hidden instance-label vectors
\(\mathbf{y}\) such that \(s = \sigma(\mathbf{y})\). Throughout the
appendices, \(c := |\mathcal{Y}|\) denotes the number of instance
classes (\(c = 10\) in the experiments; \(\mathcal{Y}\) is the instance
label set written \(Y\) in the main text). That formal
machinery is statistical: the analysis works with VC and Natarajan dimensions of
induced classifier families, Rademacher complexity of scoring classes (standard
treatments are given in~\cite{understanding-ml}), partial-risk
notation \(\mathcal{R}^{01}_{\textsf{P}}(f;\sigma)\) and the empirical
counterpart \(\widehat{\mathcal{R}}^{01}_{\textsf{P}}(f;\sigma;\mathcal{T}_{\textsf{P}})\) and weighted-model-count semantic losses. Carrying all of
this notation into the body of this paper would substantially increase the
notational burden without changing the predicate-level decomposition presented here: this
predicate-level decomposition (CP, TP and OP) operates at the logical level
where statistical-learning constants do not normally appear. A deliberately lighter
notation is therefore used in the main text, writing \(E^+(H)\text{ and }E^-(H)\) for the partition and \(\mathcal{L}_{\text{bag}}\) for the
bag-level surrogate, and confining Wang \textit{et~al.}'s \cite{wang2024learning} full statistical
apparatus to this appendix, where it is needed for the formal claims.
The correspondence is as follows.

\begin{table*}[h]
\centering\footnotesize\setlength{\tabcolsep}{5pt}
\begin{tabular}{@{}p{5.6cm}p{8.6cm}@{}}
\multicolumn{1}{@{}l}{\cellcolor{rstaclr}\tabhead{Wang {\textit{et al.}}}} &
\multicolumn{1}{l@{}}{\cellcolor{rstaclr}\tabhead{this paper}} \\
\colrule
classifier scoring function \(f \colon \mathcal{X} \to \Delta_c\)
 & probabilistic classifier \(\tilde{f}\) (CP) \\
\midrule
hard predictor \([f](x) := \arg\max_y f^y(x)\)
 & deterministic CP \(f(x)\) used in \(E^\pm\) partition (\hyperref[app:proofs]{appendix B}) \\
\midrule
transition \(\sigma \colon \mathcal{Y}^M \to \mathcal{S}\)
 & aggregation operator \(\sigma \in \Sigma\) (TP) \\
\midrule
weak observation \(s \in \mathcal{S}\)
 & bag label observed via OP \\
\midrule
zero-one risk \(\mathcal{R}^{01}(f)\)
 & expected instance misclassification rate (related to \(1 - \text{InstAcc}\)) \\
\midrule
partial risk \(\mathcal{R}^{01}_{\textsf{P}}(f;\sigma)\)
 & expected rate of \(\sigma(\mathbf{f}(\mathbf{x})) \neq s\) on hard predictions \(= 1 - \mathbb{E}[\text{Constraint}]\) \\
\midrule
empirical partial risk
\(\widehat{\mathcal{R}}^{01}_{\textsf{P}}(f;\sigma;\mathcal{T}_{\textsf{P}})\)
 & empirical rate of \(\sigma(\mathbf{f}(\mathbf{x})) \neq s\) \(= 1 - \text{Constraint}\) (computed on \(\mathcal{B}\) here) \\
\midrule
top-\(k\) partial loss \(\ell^k_\sigma(f(\mathbf{x}), s)\)
 & categorical analogue: the soft outer-product surrogate
 \(-\log \sum_{\mathbf{y} \in \sigma^{-1}(s)} \prod_i \tilde{f}(x_i)_{y_i}\) of
 \hyperref[app:wang:topk]{appendix C.5}; not literally equal to Wang \textit{et~al.}'s \cite{wang2024learning}
 Bernoulli WMC. The \(E^+/E^-\) tracking utility counts the hard
 partition (\hyperref[app:wang:topk]{appendix C.5}); a separate linear surrogate
 \(1 - \hat{p}(H, \mathbf{x})[s]\), monotone-related to but not
 identical to the \(k=1\) top-\(k\) loss, is
 disabled in all reported runs. \\
\botrule
\end{tabular}
\end{table*}

\subsection{Unambiguity conditions}
\label{app:wang:unambig}

The two conditions relied on are those introduced by Wang \textit{et~al.}\
\cite{wang2024learning}.

\begin{definition}[\(M\)-unambiguity, after Wang \textit{et~al.}\ def.~1 \cite{wang2024learning}]
\label{def:m-unambig}
\(\sigma \colon \mathcal{Y}^M \to \mathcal{S}\) is \emph{\(M\)-unambiguous}
if for any two diagonal label vectors
\(\mathbf{y} = (y, \ldots, y)\) and \(\mathbf{y}' = (y', \ldots, y')\)
with \(y \neq y'\), \(\sigma(\mathbf{y}) \neq \sigma(\mathbf{y}')\).
\end{definition}

\begin{definition}[\(1\)-unambiguity, after Wang \textit{et~al.}\ def.~2 \cite{wang2024learning}]
\label{def:1-unambig}
\(\sigma\) is \emph{\(1\)-unambiguous} if there exists a position
\(i \in \{1, \ldots, M\}\) such that, for every \(\mathbf{y} \in \mathcal{Y}^M\),
flipping \(y_i\) to any \(y_i' \neq y_i\) yields a vector
\(\mathbf{y}'\) with \(\sigma(\mathbf{y}') \neq \sigma(\mathbf{y})\).
\end{definition}

For \(M = 2\) (the bag size in the experiments) and
\(\mathcal{Y} = \{0, 1, \ldots, 9\}\), it is now verified which of the operators
satisfy each condition.

\begin{proposition}[unambiguity profile of the operators]
\label{prop:unambig-profile}
For bags of size \(M = 2\) over digit labels \(\mathcal{Y} = \{0, \ldots, 9\}\):
\begin{enumerate}
\item \(\sigma = +\) is both \(1\)-unambiguous and \(M\)-unambiguous,
\item \(\sigma = \times\) is \(M\)-unambiguous but \emph{not}
 \(1\)-unambiguous, and
\item \(\sigma = \oplus\) is \(1\)-unambiguous but \emph{not}
 \(M\)-unambiguous.
\end{enumerate}
\end{proposition}
\begin{proof}
For (i): on diagonals, \(+(y, y) = 2y\), and \(2y \neq 2y'\) for
\(y \neq y'\) in \(\mathcal{Y}\), giving \(M\)-unambiguity. For
\(1\)-unambiguity, take \(i = 1\): for any \(\mathbf{y} = (y_1, y_2)\) and
any \(y_1' \neq y_1\), \(+(y_1', y_2) - +(y_1, y_2) = y_1' - y_1 \neq 0\).
For (ii): on diagonals, \(\times(y, y) = y^2\), and \(y^2 \neq (y')^2\) for
distinct \(y, y' \in \{0, \ldots, 9\}\), giving \(M\)-unambiguity.
\(1\)-unambiguity fails: take \(\mathbf{y} = (1, 0)\) and \(y_1' = 2\);
\(\times(1, 0) = 0 = \times(2, 0)\). No choice of \(i\) works because
\(\times(y_i, 0) = 0\) for any value of \(y_i\), so position \(i\) is
uninformative whenever \(y_{\bar i} = 0\).
For (iii): on diagonals, \(\oplus(y, y) = 0\) for every \(y\), so all
diagonal vectors collapse to a single bag label, violating
\(M\)-unambiguity. \(1\)-unambiguity holds: for any
\((y_1, y_2)\) and any \(y_1' \neq y_1\),
\(\oplus(y_1', y_2) = \oplus(y_1, y_2) \oplus (y_1 \oplus y_1') \neq
\oplus(y_1, y_2)\) since \(y_1 \oplus y_1' \neq 0\).
\end{proof}

\subsection{Main learnability theorems}
\label{app:wang:theorems}

The two learnability results from Wang \textit{et~al.}\
\cite{wang2024learning} relied on are restated here, using the simplified notation of
\hyperref[app:wang:notation]{appendix C.1}.

\begin{proposition}[ERM learnability under \(M\)-unambiguity, after Wang \textit{et~al.}\ theorem~1 \cite{wang2024learning}]
\label{thm:wang-thm1}
Assume \(\sigma\) is \(M\)-unambiguous, the Natarajan dimension
\(d_{[\mathcal{F}]}\) of the induced hard-classifier class
\([\mathcal{F}]\) is finite, and the partial-risk
realizability assumption holds (there exists \(f^\star \in \mathcal{F}\) with
\(\mathcal{R}^{01}_{\textsf{P}}(f^\star; \sigma) = 0\)).
Then for any \(\epsilon, \delta \in (0, 1)\),
there is a universal constant \(C_0\) such that the empirical partial-risk
minimizer with \(\widehat{\mathcal{R}}^{01}_{\textsf{P}}(f; \sigma; \mathcal{T}_{\textsf{P}}) = 0\)
achieves classification risk \(\mathcal{R}^{01}(f) < \epsilon\) with
probability at least \(1 - \delta\), provided
\[
m_{\textsf{P}} \;\geq\; C_0 \cdot \frac{c^{2M-2}}{\epsilon^M}\,
\Bigl(d_{[\mathcal{F}]} \log(6 c M d_{[\mathcal{F}]}) \,
\log\!\bigl(c^{2M-2}/\epsilon^M\bigr)
+ \log(1/\delta)\Bigr).
\]
\end{proposition}

\begin{proposition}\label{thm:wang-prop1}
\textbf{\emph{(improved rate under joint \(1\)- and \(M\)-unambiguity, after Wang \textit{et~al.}\ \cite{wang2024learning} proposition~1)}}\, Assume \(\sigma\) is both \(1\)-unambiguous and \(M\)-unambiguous, the
Natarajan dimension \(d_{[\mathcal{F}]}\) of the induced hard-classifier
class \([\mathcal{F}]\) is finite, and
the partial-risk realizability assumption holds.
Then for any
\(\delta \in (0, 1)\) and any \(\epsilon \in (0, 1)\) that is sufficiently
close to \(0\), there is a universal constant \(C_1\) such that the
sample-complexity rate improves to
\[
m_{\textsf{P}} \;\geq\; C_1 \cdot \frac{1}{\epsilon}\,
\bigl(d_{[\mathcal{F}]} \log(6 c M d_{[\mathcal{F}]}) \log(2/\epsilon)
 + \log(1/\delta)\bigr).
\]
The smallness condition on \(\epsilon\)
(\(\epsilon \leq 1/((2M)^M c^{2M-2})\), eqn~34 there) is the regime in
which that proof controls the \(\Phi\)-transform of Wang
\textit{et~al.}'s lemma~4 \cite{wang2024learning} so that
\(\Phi_1(\epsilon/2) \leq \epsilon\); see the proposition~1 proof there.
\end{proposition}

\noindent Combining \hyperref[prop:unambig-profile]{proposition}~\ref{prop:unambig-profile} above with
\hyperref[thm:wang-thm1]{propositions}~\ref{thm:wang-thm1} and~\ref{thm:wang-prop1} of Wang \textit{et~al.}\
\cite{wang2024learning} yields the following corollary, which is the
form in which these results are used.

\begin{corollary}[predicted learnability of the operators]
\label{cor:operator-rates}
Under the framework of Wang \textit{et~al.}\ \cite{wang2024learning}, with
the finite-\(d_{[\mathcal{F}]}\) and partial-risk realizability
hypotheses of \hyperref[thm:wang-thm1]{propositions}~\ref{thm:wang-thm1}
and~\ref{thm:wang-prop1} in force:
\begin{enumerate}
\item[---] \(\sigma = +\) is learnable at the fast rate of
 \hyperref[thm:wang-prop1]{proposition}~\ref{thm:wang-prop1}, with sample complexity
 \(\widetilde{\mathcal{O}}(1/\epsilon)\) up to logarithmic factors,
\item [---]\(\sigma = \times\) is covered only by the slower guarantee of
 \hyperref[thm:wang-thm1]{proposition}~\ref{thm:wang-thm1} within this framework, with sample complexity
 \(\widetilde{\mathcal{O}}(c^{2M-2}/\epsilon^M)\), exponential in the
 bag size \(M\), and
\item [---]\(\sigma = \oplus\) admits an existential failure mode under
 Wang \textit{et~al.}'s \cite{wang2024learning} framework: the lemma~1 second clause
 \cite{wang2024learning} establishes that for non-\(M\)-unambiguous
 \(\sigma\), there exist a data distribution and a
 classifier \(f\) such that the partial risk
 \(\mathcal{R}^{01}_{\textsf{P}}(f; \sigma) = 0\), yet the classification
 risk \(\mathcal{R}^{01}(f) = 1\). This is an existential, not
 universal, statement: it does not preclude success on individual
 algorithm runs, but it implies that learnability cannot be guaranteed
 in the worst case, consistent with the bimodal seed behaviour reported empirically.
\end{enumerate}
\end{corollary}

\subsection{Implications for the empirical observations}
\label{app:wang:implications}

\hyperref[cor:operator-rates]{Corollary}~\ref{cor:operator-rates} gives a sample-complexity regime
for each operator (fast for \(+\), slow \(M\)-th-root for \(\times\) and
existential non-learnability for \(\oplus\)). These regimes predict no
gradient-descent dynamics and serve only to
\emph{interpret} the empirical behaviour reported in
\hyperref[sec:experiments-and-evaluation]{\S}{\ref{sec:experiments-and-evaluation}}, in the sense that the
observations are consistent with the regime each operator falls into.

\paragraph*{(i) Addition trains cleanly} The fast rate
\(\widetilde{\mathcal{O}}(1/\epsilon)\) of \hyperref[thm:wang-prop1]{proposition}~\ref{thm:wang-prop1}
for \(\sigma = +\) is consistent with the convergence of the \(+\)
scenario-2 runs in 15 epochs to mean bag accuracy 0.96 and mean instance
accuracy 0.98 over the 20 seeds (per-seed minima 0.93 and 0.97,
respectively).

\paragraph*{(ii) Multiplication requires careful loss balancing} The
non-\(1\)-unambiguity of \(\times\) (\hyperref[prop:unambig-profile]{proposition}~\ref{prop:unambig-profile}
case 2) is the formal condition associated with the multiplicative \emph{zero-attractor}
observed. Working at the level of the soft outer product
\(P_s = \sum_{(a, b) \in Y \times Y \,:\, a \cdot b = s} \text{cp}_1[a]\,\text{cp}_2[b]\)
(the categorical bag-marginal of \hyperref[app:wang:topk]{appendix C.5}), for every \(s > 0\) and every \(a > 0\),
\[
\partial P_s / \partial \text{cp}_1[a]
\;=\; \mathbf{1}\{a \mid s,\ s/a \in Y\} \cdot \text{cp}_2[s/a],
\]
so at \(\text{cp}_2 = \delta_0\), the entire Jacobian
\(\partial P_s / \partial \text{cp}_1[\,\cdot\,]\) is zero for every
\(s > 0\), and \(P_s\) is zero there. The log-form loss
\(-\log P_s\) (the bag cross-entropy \(\mathcal{L}_{\text{bag}}\) in the
pure-CP configuration \(w_{\text{CP}} = 1\) of
\hyperref[app:wang:topk]{appendix C.5}) is
\emph{singular} at the boundary \(P_s = 0\); the gradient of
\(-\log P_s\) is a \(0/0\) form near the attractor, and no claim is made
that it is small there (indeed along the joint perturbation
\(\text{cp}_1 = \text{cp}_2 = (1-\eta,\eta/(c-1),\ldots,\eta/(c-1))\)
its magnitude grows as \(\Theta(1/\eta)\) for small \(\eta\)). The analytic
fact relied on is the categorical-marginal one: \(P_s\) and the
Jacobian \(\partial P_s / \partial \text{cp}_1\) both \emph{saturate
towards zero} at \(\text{cp}_1 = \text{cp}_2 = \delta_0\) for every
\(s > 0\) and remain small in a neighbourhood, so the
probability mass that the constraint loss can move out of the
attractor is small there. This probability-mass saturation, together
with the log-loss singularity at \(P_s = 0\), is why \(\sigma = \times\)
is sensitive to loss balancing: as documented in
\hyperref[sec:experiments-and-evaluation]{\S}{\ref{sec:experiments-and-evaluation}} (experiment~1), the
four-term loss with auxiliary entropy and ILP-style regularizers
disabled (\(\lambda_{\text{ent}} = \lambda_{\text{ilp}} = 0\)) avoids
the attractor, while adding those auxiliaries with non-zero weights
(\(\lambda_{\text{ent}} = 0.15\text{ and } \lambda_{\text{ilp}} = 0.5\)) drives
the optimizer into it. \hyperref[thm:wang-thm1]{Proposition}~\ref{thm:wang-thm1}
of Wang \textit{et~al.}\ \cite{wang2024learning} guarantees that learning is
still possible, but only at the slower \(M\)-th-root sample-complexity
rate. The theorem is a sample-complexity statement and does not
predict gradient-descent dynamics; the empirical observation
of higher per-seed variance and the practical need for a carefully
balanced loss (the four-term configuration of
\hyperref[alg:unified]{algorithm}~\ref{alg:unified} with weights
\((\lambda_{\text{bag}}, \lambda_{\text{con}}, \lambda_{\text{cp}},
\lambda_{\text{ctp}}) = (1.0, 1.0, 0.05, 0.5)\), rather than an
alternative variant with \(\lambda_{\text{ent}} = 0.15, \lambda_{\text{ilp}} = 0.5\))
is \emph{consistent with} the slower-rate regime predicted by
\hyperref[thm:wang-thm1]{proposition}~\ref{thm:wang-thm1}, as reported in
\hyperref[sec:experiments-and-evaluation]{\S}{\ref{sec:experiments-and-evaluation}}.

\paragraph*{(iii) XOR exhibits the non-\(M\)-unambiguous failure mode} The
XOR scenario-2 runs achieve bag accuracy and constraint satisfaction
both approximately \(0.96\) across all 20 seeds, but display a
\emph{bimodal} instance accuracy: roughly half of the seeds converge to
instance accuracy above \(0.9\), and the remaining seeds collapse to
instance accuracy near \(0\), yielding the aggregate
\(0.439 \pm 0.497\) reported in \hyperref[tab:exp2_cp]{table}~\ref{tab:exp2_cp}. A canonical
mechanism explaining this pattern is an \emph{XOR-equivariant
relabelling}: any classifier \(\tilde{f}\) satisfying
\(\tilde{f}(x_1) \oplus \tilde{f}(x_2) = y_1 \oplus y_2 = s\) on every
bag preserves bag accuracy and constraint satisfaction regardless of
how it maps individual instances. The canonical parametrized family
is \(\tilde{f}(x) = y(x) \oplus \kappa\) for a fixed constant \(\kappa\)
that is label-preserving on the digit range \(\mathcal{Y} = \{0, \ldots, 9\}\),
i.e. \(y \oplus \kappa \in \mathcal{Y}\) for every \(y \in \mathcal{Y}\);
the only such \(\kappa\) are \(0\) and \(1\) (any larger \(\kappa\) takes
some digit out of \(\mathcal{Y}\)). On this family,
\((y_1 \oplus \kappa) \oplus (y_2 \oplus \kappa) = y_1 \oplus y_2 = s\),
so bag and constraint accuracy are preserved; instance accuracy is
correct at \(\kappa = 0\) and zero at \(\kappa = 1\). The \(\kappa = 1\)
realization is the parity flip \(0 \leftrightarrow 1,
2 \leftrightarrow 3, \ldots, 8 \leftrightarrow 9\). The collapsed seeds are characterized through this family at the level of the bag identity it
satisfies; the seed-level confusion matrix is not separately inspected
to identify which specific relabelling each collapsed seed realizes.

This pattern is \emph{consistent with} the existential failure mode
captured by Wang \textit{et~al.}'s lemma~1 \cite{wang2024learning}: the lemma
proves the existence of a data distribution and a
classifier \(f\) with \(\mathcal{R}^{01}_{\textsf{P}}(f; \oplus) = 0\)
yet \(\mathcal{R}^{01}(f) = 1\). Any non-trivial XOR-equivariant
relabelling \(\tilde{f}(x) = y(x) \oplus \kappa\) with \(\kappa \neq 0\) would
realize this configuration on the actual MNIST distribution: partial risk is zero
because every bag XOR is preserved, and instance risk is one because every
instance label is wrong. The connection to lemma~1 is that both
phenomena exploit the non-\(M\)-unambiguity of \(\oplus\) (the
diagonal-vector collapse \(\oplus(y, y) = 0\) implies that
XOR-equivariant relabellings form a non-trivial isotropy group for
\(\sigma\)); the formal statement is existential, and the empirical
observation is consistent with a gradient-descent attractor in that
isotropy group.
The class-specific bound of Tsamoura
\textit{et~al.}~\cite{tsamoura2025imbalances} (proposition~3.1)
permits the uniform-bad per-class attainment that such a relabelling
would realize, without predicting it; the point is developed in
\hyperref[app:tsamoura:implications]{appendix D.4}.

\subsection{Top-\(k\) surrogate and bag-level cross-entropy}
\label{app:wang:topk}

Wang \textit{et~al.}\ \cite{wang2024learning} def.~3 introduces the
\emph{top-\(k\) partial loss}
\begin{equation}
\ell^k_\sigma(f(\mathbf{x}), s) \;:=\; \mathrm{SL}\Bigl(\bigvee_{i = 1}^{k}
 \mathbf{y}^{(i)},\; f(\mathbf{x})\Bigr),
\label{eq:wang-topk}
\end{equation}
where \(\mathbf{y}^{(1)}, \ldots, \mathbf{y}^{(k)}\) are the top-\(k\)
maximizers of \(P_{f(\mathbf{x})}(\mathbf{y}) := \prod_i \tilde{f}(x_i)_{y_i}\)
in the pre-image \(\sigma^{-1}(s)\), and \(\mathrm{SL}\) is the semantic
loss (negative log weighted-model-count).

For \(k = |\sigma^{-1}(s)|\), the disjunction is a positive DNF (Disjunctive Normal Form) formula
over Boolean variables \(A_{i, y}\) (true iff instance \(i\) is assigned
label \(y\)). In Wang \textit{et~al.}'s \cite{wang2024learning} WMC semantics, the Boolean variables are
\emph{\textit{a priori}} independent, so different label-vector conjunctions
\(\boldsymbol{\Phi}_{\mathbf{y}^{(i)}}\) and \(\boldsymbol{\Phi}_{\mathbf{y}^{(j)}}\)
can be simultaneously satisfied by Boolean interpretations that assign
each instance multiple labels. Adding an exclusiveness constraint
\(\Xi\) of the form \(\neg(A_{i, y} \wedge A_{i, y'})\) for every
\(i\) and every \(y \neq y'\) reduces the model count by suppressing
those multi-label interpretations (Wang \textit{et~al.}\
\cite{wang2024learning}~\S A.1, discussion following lemma~2).

The framework here instead uses a \emph{categorical} model directly: each
\(\tilde{f}(x_i)\) is a softmax over \(Y\) summing to one. The
quantity
\(\sum_{\mathbf{y} \in \sigma^{-1}(s)} \prod_i \tilde{f}(x_i)_{y_i}\)
computed as the soft outer product is the marginal probability
\(\mathbb{P}(\sigma(\mathbf{y}) = s)\) under independent categorical
draws \(y_i \sim \tilde{f}(x_i)\). This is treated as a categorical
analogue of Wang \textit{et~al.}'s \cite{wang2024learning} WMC: the two objects sum over the same set
of witnessing label-vector assignments in the pre-image,
but the underlying probabilistic semantics are different
(Wang \textit{et~al.}'s \cite{wang2024learning} WMC sums over independent-Bernoulli interpretations,
each contributing both true-variable factors \(\omega(A_{i,y})\) and
false-variable factors \(1 - \omega(A_{i,y'})\); the categorical
product has no false-variable factors). The two formulas are not
identical in general and coincide only on degenerate inputs (e.g.\
one-hot \(\tilde{f}(x_i)\)). The categorical form is used throughout
because it is the standard output of a neural classifier.

Under this categorical reading, when the bag prediction
\(\hat{s}(H, \mathbf{x})\) is computed purely from CP through the
soft outer product (\hyperref[alg:unified]{algorithm}~\ref{alg:unified} mixing weights
\(w_{\text{CP}} = 1\text{ and } w_{\text{OP}} = w_{\text{TP}} = 0\)), the
bag-level cross-entropy loss
\(\mathcal{L}_{\text{bag}} = -\log \hat{p}(H, \mathbf{x})[s]\)
takes the form
\(-\log \sum_{\mathbf{y} \in \sigma^{-1}(s)} \prod_i \tilde{f}(x_i)_{y_i}\),
i.e.\ the categorical-marginal counterpart of Wang \textit{et~al.}'s
\cite{wang2024learning} full-pre-image top-\(k\) loss for \(k = |\sigma^{-1}(s)|\). For the
more general case \(w_{\text{CP}} \in (0, 1)\), the combined bag
prediction \(\hat{p}\) mixes \(\hat{p}_{\text{CP}}\) with
\(p_{\text{OP}}\) and \(\hat{p}_{\text{TP}}\); only the
\(\hat{p}_{\text{CP}}\) component is the categorical analogue of
Wang \textit{et~al.}'s \cite{wang2024learning} top-\(k\)/WMC family. Wang \textit{et~al.}'s theorem~2 \cite{wang2024learning} is stated
for the WMC/top-\(k\) loss with Bernoulli-variable semantics; it is
not stated for the categorical bag cross-entropy here, and no Rademacher
argument is provided that rederives the bound for the
categorical surrogate. The implication drawn below is therefore by
analogy with Wang \textit{et~al.}'s \cite{wang2024learning} formal result rather than as a direct
corollary; the mixed training objective of
\hyperref[alg:unified]{algorithm}~\ref{alg:unified}, with the additional structural terms
\(\mathcal{L}_{\text{con}}, \mathcal{L}_{\text{cp}}\text{ and } \mathcal{L}_{\text{ctp}}\),
is an empirical extension that no Wang \textit{et~al.}\ \cite{wang2024learning} bound covers in either
literal or analogue form.

Theorem~2 of Wang \textit{et~al.}\ \cite{wang2024learning} requires \(\sigma\)
to be \emph{both \(1\)- and \(M\)-unambiguous}, and yields, for any
integer \(k \geq 1\) and \(\delta \in (0, 1)\), with probability at
least \(1 - \delta\), the
following Rademacher-style bound on the classification risk:
\begin{equation}
\mathcal{R}^{01}(f) \;\leq\; \Phi\!\left( (k + 1)\,\left(
\widehat{\mathcal{R}}^{k}_{\textsf{P}}(f; \sigma; \mathcal{T}_{\textsf{P}})
+ 2\sqrt{2 k}\, M^{3/2} \mathfrak{R}_{M m_{\textsf{P}}}(\mathcal{F})
+ \sqrt{\tfrac{\log(1/\delta)}{2 m_{\textsf{P}}}}\right)\right),
\label{eq:wang-thm2-bound}
\end{equation}
where \(\Phi\) is an increasing function with \(\Phi(t)/t \to 1\) as \(t \to 0\), and
\(\mathfrak{R}_m(\mathcal{F})\) denotes the Rademacher complexity of the
scoring class \(\mathcal{F}\)~\cite{Rademacher}. By
\hyperref[prop:unambig-profile]{proposition}~\ref{prop:unambig-profile}, the joint
\(1\)- and \(M\)-unambiguity requirement is satisfied only by
\(\sigma = +\) among the operators studied here; \(\sigma = \times\) is
covered only by \hyperref[thm:wang-thm1]{proposition}~\ref{thm:wang-thm1}'s slower
\(M\)-unambiguous rate, and \(\sigma = \oplus\) is not covered by
either aggregate Wang \textit{et~al.}\ \cite{wang2024learning} bound.
When the bag prediction is computed from CP through the soft outer
product with \(k = |\sigma^{-1}(s)|\) (equivalently, the
\hyperref[alg:unified]{algorithm}~\ref{alg:unified} configuration
\(w_{\text{CP}} = 1\text{ and } w_{\text{OP}} = w_{\text{TP}} = 0\)), the bag-CE
term \(\mathcal{L}_{\text{bag}}\) is the categorical analogue of
Wang \textit{et~al.}'s \cite{wang2024learning} full-pre-image top-\(k\) loss. For \(\sigma = +\), which
is both \(1\)- and \(M\)-unambiguous, \linkref[equation]{eq:wang-thm2-bound}
\emph{as stated} applies to Wang \textit{et~al.}'s \cite{wang2024learning} WMC/top-\(k\) surrogate;
it is used only as an analogue for the categorical bag-CE, since
no separate Rademacher derivation is given here that would extend
the bound literally to \(\mathcal{L}_{\text{bag}}\). With that
analogue reading, the bound gives a high-probability upper bound on
the classification risk \(\mathcal{R}^{01}(f)\) in terms of the
\emph{empirical} top-\(k\) risk and a Rademacher-plus-Hoeffding
deviation term, and complements the deterministic
\hyperref[prop:surrogate]{proposition}~\ref{prop:surrogate}. The bound is conditional in the
standard sense (a small empirical top-\(k\) risk plus a small
deviation term yields a small \(\mathcal{R}^{01}(f)\)), and does
not guarantee that gradient descent achieves the
small-empirical-risk precondition. \hyperref[prop:surrogate]{Proposition}~\ref{prop:surrogate}
provides the complementary deterministic statement
\(|E^-| \leq \mathcal{L}_{\text{bag}} / \log 2\), so
\(\mathcal{L}_{\text{bag}} / \log 2\) upper-bounds the discrete
misclassification count.

Note that the $E^+/E^-$ tracking routine is a
\emph{tracking-only utility}, used to produce the per-epoch hard
partition counts of \hyperref[fig:eplus_eminus]{figure}~\ref{fig:eplus_eminus}
(\(\sigma\) applied to the hard per-instance predictions, counted
against \(s_b\) on the validation set, per candidate operator); it does
\emph{not} contribute to the training loss. A separate linear-surrogate
loss \(1 - \hat{p}(H, \mathbf{x})[s]\), monotone-related to the
log-form top-\(k\) loss of Wang \textit{et~al.}\ \cite{wang2024learning} but not formally identical,
exists in the implementation and is disabled (weight zero) in all
reported runs. The
training loss used throughout the experiments is the four-term
\(\mathcal{L}_{\text{bag}} + \lambda_{\text{con}} \mathcal{L}_{\text{con}}
+ \lambda_{\text{cp}} \mathcal{L}_{\text{cp}}
+ \lambda_{\text{ctp}} \mathcal{L}_{\text{ctp}}\) of
\hyperref[alg:unified]{algorithm}~\ref{alg:unified}; only the bag-CE component
\(\mathcal{L}_{\text{bag}}\) is the component analogous to the
Wang \textit{et~al.}\ top-\(k\)/WMC family in
\linkref[equation]{eq:wang-thm2-bound}. The three structural terms
\(\mathcal{L}_{\text{con}}, \mathcal{L}_{\text{cp}}\text{ and } \mathcal{L}_{\text{ctp}}\)
provide additional inductive bias on top of that surrogate; bounding
the population risk under the full four-term loss would require a
separate Rademacher argument and is beyond the scope of this paper.

\subsection{Background knowledge and unknown transitions}
\label{app:wang:bk}

When \(\sigma\) is unknown, Wang \textit{et~al.}\ \cite{wang2024learning} consider
a \emph{transition class} \(\mathcal{G}\) of candidate transitions
including the true one. Wang \textit{et~al.}'s def.~5 introduces the
\emph{unambiguity of \(\mathcal{G}\)} condition: for each
\(\sigma' \in \mathcal{G}\), each diagonal label vector
\(\mathbf{y} = (l_i, \ldots, l_i)\) and each label \(l_j \neq l_i\),
some vector \(\mathbf{y}' \in \{l_i, l_j\}^M\) satisfies
\(\sigma'(\mathbf{y}') \neq \sigma(\mathbf{y})\), where \(\sigma\) is the
true transition. A scoring class
\(\mathcal{F}\) is \emph{\(r\)-bounded} (Wang \textit{et~al.}'s~\S5) if there exists
\(r > 0\) such that \(\mathbb{P}(\,[f](X) = y \mid Y = y\,) \geq r\) for
every \(y \in \mathcal{Y}\) and every \(f \in \mathcal{F}\); i.e. every
classifier in the search space has class-conditional hard-prediction
accuracy at least \(r\) on every class. Under the unambiguity of \(\mathcal{G}\) and the
\(r\)-boundedness of \(\mathcal{F}\), theorem~5 of Wang \textit{et~al.}\
\cite{wang2024learning} establishes
\(\mathcal{R}^{01}(f) \leq \mathcal{O}(\mathcal{R}^{01}_{\textsf{P}}(f; \mathcal{G})^{1/M})\)
as \(\mathcal{R}^{01}_{\textsf{P}}(f; \mathcal{G}) \to 0\), with sample complexity
\begin{align*}
m_{\textsf{P}} \geq{} & C_4 \cdot \frac{c^{2M-2}}{r^M \epsilon^M} \\
& \cdot \Bigl(\bigl((d_{[\mathcal{F}]} + d_{\mathcal{G}})\log(6M(d_{[\mathcal{F}]} + d_{\mathcal{G}}))
+ d_{[\mathcal{F}]}\log c\bigr) \\
& \quad \cdot \log\!\bigl(c^{2M-2}/(r^M \epsilon^M)\bigr)
+ \log(1/\delta)\Bigr).
\end{align*}
The key new term is \(d_{\mathcal{G}}\), the VC dimension of the family of
indicator functions
\(\{(\mathbf{y}, s) \mapsto \mathbb{1}\{\sigma'(\mathbf{y}) \neq s\} :
\sigma' \in \mathcal{G}\}\). The \emph{background-knowledge size study}
experiment, in which the candidate-operator set \(\Sigma\), the
ILP-style realization of Wang \textit{et~al.}'s \cite{wang2024learning} transition class \(\mathcal{G}\),
is varied from \(|\Sigma| = 1\) to \(|\Sigma| = 5\), corresponds
operationally to varying \(d_{\mathcal{G}}\). The monotonic increase in
convergence epoch reported here (\hyperref[app:bk_sweep]{appendix A.2},
\hyperref[tab:bk_sweep]{table} {\ref{tab:bk_sweep}}) is the empirical \emph{analogue} of the
qualitative direction of Wang \textit{et~al.}'s theorem~5\cite{wang2024learning}: the bound is on
sample complexity \(m_{\textsf{P}}\) and the quantity here is on optimization
epochs, so the two quantities are not the same, and no claim is made
to test theorem~5 directly. The observed monotonicity in epoch count
is consistent with the qualitative direction of the bound (more
candidates require more data/effort) and serves as an empirical
illustration rather than a formal corroboration.

\subsection{What is adopted and what is added}
\label{app:wang:subsumes}

The following are taken directly from Wang \textit{et~al.}\ \cite{wang2024learning}
and used to ground the analysis here:
\begin{enumerate}
\item[---] A formal condition, failure of \(1\)-unambiguity
 (def.~\ref{def:1-unambig}), with which the multiplicative
 zero-attractor observed here is consistent.
\item[---] A formal condition consistent with the high-bag/low-instance
 failure mode of XOR (failure of \(M\)-unambiguity, together with the
 existential second clause of Wang \textit{et~al.}\ \cite{wang2024learning} lemma~1).
\item[---] A probabilistic Rademacher-style risk bound on the
 top-\(k\)/WMC surrogate (theorem~2 of Wang \textit{et~al.}\ \cite{wang2024learning}, restated as
 \linkref[equation]{eq:wang-thm2-bound}), imported \emph{by
 analogy} onto the pure-CP categorical surrogate; no separate
 Rademacher argument is given for the categorical form, and the
 bound is not claimed to cover the mixed four-term loss here.
\item[---] A theoretical anchor for the warm-CP requirement (\(r\)-boundedness
 in theorem~5 of Wang \textit{et~al.}\ \cite{wang2024learning}).
\end{enumerate}

The contributions of the present paper, built on top of the above, are:
\begin{enumerate}
\item[---] A \emph{structured} decomposition of the hypothesis into the three
 predicates CP, TP and OP, in place of a monolithic scoring function. This
 decomposition is what makes the Bag--Constraint and Bag--Inst gap
 metrics, and consequently the reasoning-shortcut diagnostics of
 \cite{marconato23} and \cite{bears2024}, well-defined within an ILP setting.
\item[---] A diagnostic apparatus that separates distinct failure
 patterns: the ablations `no CP', `no TP' and `OP only' in
 \hyperref[tab:unified_ablation]{table}~\ref{tab:unified_ablation} expose the CP-shortcut pattern
 (`no CP') and the no-signal collapse (`no TP' and `OP only', which are
 near-indistinguishable) that a risk-only analysis does not separate.
\item[---] An empirical probe of Wang \textit{et~al.}'s \cite{wang2024learning} transition-class term
 \(d_{\mathcal{G}}\) via the background-knowledge size study, which
 varies the candidate operator class and illustrates the corresponding
 effect on optimization convergence speed, without claiming to measure
 \(d_{\mathcal{G}}\) or to test the sample-complexity bound directly.
\item[---] A formalization of reasoning shortcuts via the symbolic
 \(\mathbb{I}_\sigma\) structures
 (\hyperref[sec:problem-structure-and-predicate-formulation]{\S}\ref{sec:problem-structure-and-predicate-formulation}), which
 permits explicit counterexamples, such as the XOR-equivariant
 relabelling family of \hyperref[app:wang:implications]{appendix C.4},
 to be constructed without recourse to risk bounds.
\end{enumerate}

\renewcommand{\thesection}{D}
\setcounter{section}{0}
\section*{Appendix D. Mapping to class-specific imbalance theory}
\label{app:tsamoura}

The recent work of Tsamoura \textit{et~al.}~\cite{tsamoura2025imbalances}
extends the MI-PLL learnability framework of Wang \textit{et~al.}~\cite{wang2024learning}
in three directions that bear directly on the empirical findings here: (i)
Tsamoura \textit{et~al.}\ bound the \emph{class-specific} risk \(R_j(f) := \mathbb{P}([f](X) \neq j \mid Y = j)\)
rather than only the aggregate risk \(R(f)\), (ii) those bounds do not
require \(M\)-unambiguity, and a relaxed form recovers Wang \textit{et~al.}'s lemma~1 \cite{wang2024learning} in the
\(M\)-unambiguous \(M = 2\) case, and (iii) Tsamoura \textit{et~al.}\ propose mitigation algorithms
that operate at training and testing time. This appendix supplies the
correspondence between their framework and the empirical patterns
reported in \hyperref[sec:experiments-and-evaluation]{\S}\ref{sec:experiments-and-evaluation}.

\setcounter{subsection}{0}
\subsection{Notation}
\label{app:tsamoura:notation}

Tsamoura \textit{et~al.} \cite{tsamoura2025imbalances} refer to the MI-PLL setting as NeSy. That sample
structure is the same as Wang \textit{et~al.}'s \cite{wang2024learning} and that of the present work: training samples
\((\mathbf{x}, s)\) with \(\mathbf{x} = (x_1, \ldots, x_M)\), hidden
gold labels \(\mathbf{y} = (y_1, \ldots, y_M)\), and weak label
\(s = \sigma(\mathbf{y})\). That general definition does not require
the \(x_i\) within a bag to be i.i.d. Their~\S3 theory
(propositions~3.1, 3.3 and 3.4), however, does assume i.i.d. on
\(\mathcal{D}_{\textsf{P}}\) and focuses on \(M = 2\) to simplify
presentation; both the theory and the~\S4 mitigation
algorithms operate under known \(\sigma\). These results are imported into the present framework only for the i.i.d., \(M = 2\),
known-\(\sigma\) scenario-2 setting. Tsamoura \textit{et~al.}'s \cite{tsamoura2025imbalances} lighter
typography is followed in this appendix: \(R_j(f)\) and \(R_{\textsf{P}}(f; \sigma)\)
denote the same per-class and partial zero-one risks that
\hyperref[app:wang]{appendix}~\ref{app:wang} writes as \(\mathcal{R}^{01}_j(f)\) and
\(\mathcal{R}^{01}_{\textsf{P}}(f; \sigma)\), respectively. The quantity
\(R_j(f)\) corresponds to the per-class instance misclassification rate
(related to \(1 - \mathrm{InstAcc}\) restricted to the gold class \(j\)),
and \(R_{\textsf{P}}(f; \sigma)\) is Wang \textit{et~al.}'s \cite{wang2024learning} partial risk,
corresponding to \(1 - \mathbb{E}[\mathrm{Constraint}]\) here
(\hyperref[app:wang:notation]{appendix C.1}); for a deterministic
classifier it also matches \(|E^-(H)|/N\) in expectation.

\subsection{Class-specific risk bounds}
\label{app:tsamoura:thm}

Proposition~3.1 of Tsamoura \textit{et~al.}\ establishes that, for any \(j \in \mathcal{Y}\),
\[
R_j(f) \;\leq\; \Phi_{\sigma, j}\!\bigl(R_{\textsf{P}}(f; \sigma)\bigr),
\]
where \(\Phi_{\sigma, j}\) is the optimal value of a quadratically
constrained program with a linear objective \(\mathbf{w}_j^\top \mathbf{h}\)
(eqn\ 2 of Tsamoura \textit{et~al.}), depending jointly on \(\sigma\), the label-marginal
vector \(\mathbf{r}\) and the column index \(j\). The single quadratic
constraint \(\mathbf{h}^\top \boldsymbol{\Sigma}_{\sigma, \mathbf{r}} \mathbf{h}
= R_{\textsf{P}}(f; \sigma)\) pins the program to the
partial-risk level set; the remaining constraints are linear. The shape of
\(\Phi_{\sigma, j}\) is what controls the per-class learning
imbalance: classes for which \(\Phi_{\sigma, j}\) rises steeply with
the partial risk are intrinsically harder to learn even when the
hidden marginals are uniform.

\subsection{Recovery of the lemma~1 bound of Wang \textit{et~al.}}
\label{app:tsamoura:wang-recovery}

Proposition 3.4 of Tsamoura \textit{et~al.}\ shows that, if \(\sigma\) is \(M\)-unambiguous,
relaxing the quadratically constrained
program in their eqn~2 (dropping the
positivity and normalization constraints, restricting
\(\boldsymbol{\Sigma}_{\sigma, \mathbf{r}}\) to its diagonal and replacing
the equality with an inequality) recovers Wang \textit{et~al.}'s
lemma~1\ \cite{wang2024learning}: \(R(f) \leq \sqrt{c(c-1) \, R_{\textsf{P}}(f; \sigma)}\) for
\(M = 2\). Hence, in the experiments here
(\hyperref[sec:experiments-and-evaluation]{\S}\ref{sec:experiments-and-evaluation}), the Wang-anchored
sample-complexity statements of \hyperref[app:wang:theorems]{appendix C.3}
apply only to the \(M\)-unambiguous operators in the present set, namely
\(\sigma = +\) (the fast-rate \hyperref[thm:wang-prop1]{proposition}~\ref{thm:wang-prop1}, which also
requires \(1\)-unambiguity) and \(\sigma = \times\) (the slower
\(M\)-th-root rate of \hyperref[thm:wang-thm1]{proposition}~\ref{thm:wang-thm1}); these statements do
\emph{not} apply to \(\sigma = \oplus\), which fails
\(M\)-unambiguity and falls outside the scope of both Wang
aggregate-rate theorems. Tsamoura \textit{et~al.}'s \cite{tsamoura2025imbalances} framework supplies a class-specific bound for
\(\sigma = \oplus\) that drops the \(M\)-unambiguity requirement
(a strict generalization of Wang \textit{et~al.}'s lemma~1 \cite{wang2024learning} via Tsamoura \textit{et~al.}'s
proposition~3.4), without making aggregate sample-complexity
predictions for that operator.

\subsection{Implications for the empirical observations}
\label{app:tsamoura:implications}

\paragraph*{Why multiplication tolerates high OP noise (\hyperref[sec:experiments-and-evaluation]{\S}\ref{sec:experiments-and-evaluation}, \hyperref[tab:op_noise]{table}~\ref{tab:op_noise}, \hyperref[fig:op_noise]{figure} \ref{fig:op_noise})}
The finding that \(\sigma = \times\) remains at or above
\(\mathrm{ConsRatio} \geq 0.83\) up to noise level \(0.7\) while
\(\sigma \in \{+, \oplus\}\) drop sharply between \(0.5\) and \(0.7\) is
empirically related to the pre-image-asymmetry mechanism that Tsamoura
\textit{et~al.}\ \cite{tsamoura2025imbalances} describe under clean weak-label distributions. The
illustrative example (figure~2 of Tsamoura \textit{et~al.}\ \cite{tsamoura2025imbalances}) is
\(\sigma = \mathrm{max}\) on MNIST digits, where the pre-image of
\(s = 0\) is the \emph{smallest} (only the diagonal pair \((0, 0)\)),
which makes \(s = 0\) the rarest weak label under a uniform digit marginal
(probability \(1/100\)) and produces the steepest \(\Phi_{\sigma, 0}\)
curve in the class-specific bound, so class \(0\) is hardest to learn.
For \(\sigma = \times\) on the digit range, the pre-image of \(s = 0\)
is instead the \emph{largest} (the zero-absorber
\(\{(0, y), (y, 0) : y \in \mathcal{Y}\}\), \(19\) ordered pairs out of
\(100\)). Thus, whenever corruption produces or leaves a zero bag label,
the multiplicative rule has many instance-level explanations available;
this helps explain why zero-absorbing multiplication can be empirically
more tolerant to OP corruption than addition or XOR. Note that Tsamoura \textit{et~al.}'s \cite{tsamoura2025imbalances} theory
does not directly model OP corruption: the bounds are stated for a clean
weak-label structure and known \(\sigma\). The OP-noise plateau is
therefore an empirical observation with a plausible structural explanation:
pre-image asymmetry at \(s = 0\), which is related to the mechanism that Tsamoura
\textit{et~al.}\ \cite{tsamoura2025imbalances} identify for clean weak-label imbalances. The direction of the asymmetry differs:
small for \(\mathrm{max}\) (class \(0\) hardest to learn under clean data)
and large for \(\times\) (\(s = 0\) most noise-tolerant under corrupted
data); Tsamoura \textit{et~al.}'s \cite{tsamoura2025imbalances} framework gives the structural language but not
a formal bound for the noise regime observed here.

\paragraph*{XOR's bimodal instance recovery
(\hyperref[sec:experiments-and-evaluation]{\S}\ref{sec:experiments-and-evaluation}, \hyperref[tab:exp2_cp]{table}~\ref{tab:exp2_cp})}The finding that approximately half the seeds show the
bag-accurate/instance-collapsed pattern when \(\sigma = \oplus\)
(\hyperref[app:wang:implications]{appendix C.4}) is consistent with collapse to
an XOR-equivariant relabelling \(\tilde{f}(x) = y(x) \oplus \kappa\)
for some \(\kappa \neq 0\); such a relabelling would realize the lemma~1 configuration of Wang \textit{et~al.}\ \cite{wang2024learning} on the MNIST
distribution (zero partial risk because the bag XOR is preserved by
the relabelling, instance risk one because every instance label is
wrong). Under this hypothesized mechanism, the per-class risk vector
would be uniform (\(R_j(\tilde{f}) = 1\) for every \(j\), since each
digit is mapped to the XOR-shifted partner), rather than the maximally
imbalanced single-class-zero profile that Tsamoura \textit{et~al.}'s
proposition~3.1 \cite{tsamoura2025imbalances} worst case considers. Tsamoura \textit{et~al.}'s proposition~3.1
\cite{tsamoura2025imbalances} bound \(\Phi_{\sigma, j}\) permits this uniform-bad attainment but does
not predict it; both the worst-case imbalanced attainment and the
uniform attainment lie under the upper envelope, and the bound says
nothing about which one gradient descent realizes.

\subsection{Mitigation algorithms}
\label{app:tsamoura:mitigation}

Tsamoura \textit{et~al.} \cite{tsamoura2025imbalances} propose three concrete algorithms: a \MakeLowercase{Label Ratio
Solver} (algorithm~1 there) for estimating the hidden-label marginal
\(\hat{\mathbf{r}}\) from weak labels alone; an LP
training-time pseudolabel assignment (\S4.2 there) that adheres
to both the classifier scores and \(\hat{\mathbf{r}}\), and the
test-time CAROT procedure (algorithm~2 there) that adjusts model
scores via a robust semi-constrained optimal transport. That
empirical evaluation reports substantial accuracy improvements on
NeSy benchmarks over semantic-loss and long-tailed-learning baselines. Within the CP/TP/OP
decomposition, these three algorithms map to:

\begin{enumerate}
\item[---] The marginal estimation supplies an input to the CP predicate:
 an estimate of \(\mathbb{P}(Y = j)\) for each digit class \(j\),
 which the framework currently treats as uniform.
\item[---] The LP-based pseudolabels can replace the soft outer product
 used in the \(\mathcal{L}_{\mathrm{ctp}}\) term: each batch's
 pseudolabels \(Q_i\) become CP targets that adhere to \(\sigma\) (TP)
 and \(\hat{\mathbf{r}}\), enforcing the OP constraint at the level
 of explicit Boolean conjunctions rather than through the surrogate
 cross-entropy.
\item[---] CAROT acts at test time on the CP outputs to enforce marginal
 consistency without retraining; this is applicable when the
 training-time CP/TP/OP loop has already converged to a class-imbalanced solution.
\end{enumerate}

Combining Tsamoura \textit{et~al.}'s \cite{tsamoura2025imbalances} imbalance-mitigation algorithms with the
ILP-induced CP/TP/OP semantics of \hyperref[sec:problem-structure-and-predicate-formulation]{\S2b}
is a direct extension left for future work: it could help address the
XOR bimodal collapse reported here and the post-plateau collapse in
multiplication under OP noise.

\end{document}